\documentclass{statsoc}

\usepackage[a4paper]{geometry}
\usepackage{graphicx}
\usepackage{amssymb}
\usepackage{amsmath}
\usepackage{natbib}
\usepackage{color}
\usepackage{float}
\usepackage{subfigure}
\usepackage{algorithm}
\usepackage{bm}
\usepackage[noend]{algpseudocode}
\usepackage{lscape}
 \usepackage{color}
\usepackage[colorinlistoftodos]{todonotes}

\usepackage{multirow}
\usepackage{float}
\def\t{\textrm}
\def\d{\partial}
\def\mb{\mathbf}
\def\bb{\mathbb}
\def\R{\mathbb{R}}

\newtheorem{theorem}{Theorem}[section]

\newtheorem{remark}[theorem]{Remark}

\title[In-GP on manifolds]{Intrinsic Gaussian processes on complex constrained \\ domains}
\author[Mu Niu {\it et al.}]{Mu Niu}
\address{ Centre for Mathematical Sciences, School of Computing, Electronics and Mathematics, Plymouth University, Plymouth, UK.}
\email{mu.niu@plymouth.ac.uk }
\author{Pokman Cheung}
\address{}\email{pokman@alumni.stanford.edu}
\author{Lizhen Lin}
\address{Department of Applied and Computational Mathematics and Statistics, The University of Notre Dame,  USA.}
\email{lizhen.lin@nd.edu}
\author{Zhenwen Dai}
\author[Mu Niu {\it et al.}]{Neil Lawrence}
\address{The University of Sheffield and Amazon.com, UK.}
\email{zhenwend@amazon.com,\; lawrennd@amazon.co.uk }
\author[Mu Niu {\it et al.}]{David Dunson}
\address{Department of Statistical Science, Duke University}
\email{dunson@duke.edu}

\begin{document}
\begin{abstract}
We propose a class of intrinsic Gaussian processes (in-GPs) for interpolation, regression and classification on manifolds with a primary focus on complex constrained domains or irregular-shaped spaces arising as subsets or submanifolds of $\R$, $\R^2$, $\R^3$ and beyond.  For example, in-GPs can accommodate spatial domains arising as complex subsets of Euclidean space.  in-GPs respect the potentially complex boundary or interior conditions as well as the intrinsic geometry of the spaces.  The key novelty of the proposed approach is to utilise the relationship between heat kernels and the transition density of Brownian motion on manifolds for constructing and approximating valid and computationally feasible covariance kernels.  This enables in-GPs to be practically applied in great generality, while existing approaches for smoothing on constrained domains are limited to simple special cases.  The broad utilities of the in-GP approach is illustrated through simulation studies and data examples.
\end{abstract}

\section{Introduction}\label{intro}

\begin{figure}
\centering 
\includegraphics[width=4.6cm]{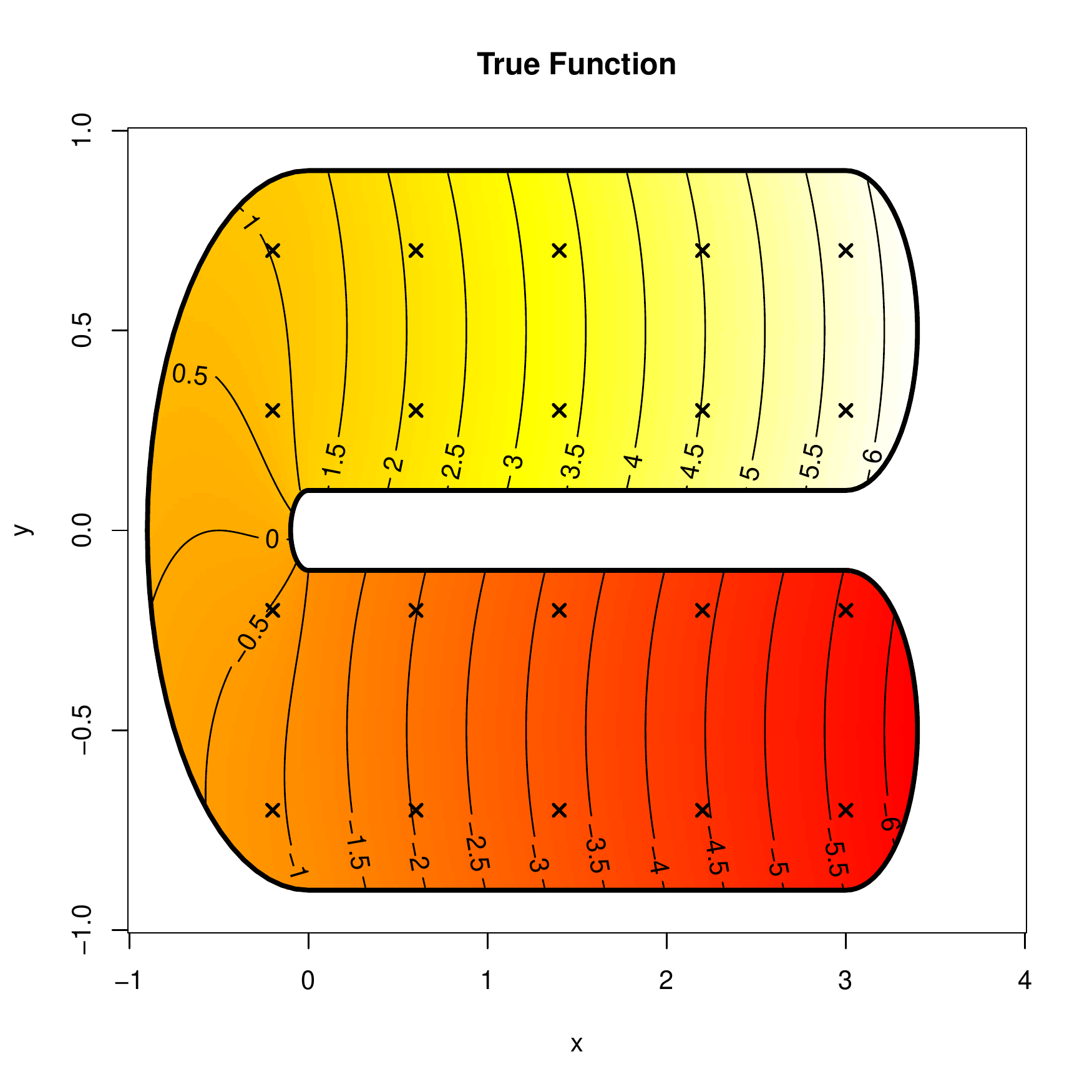}
\includegraphics[width=4.6cm]{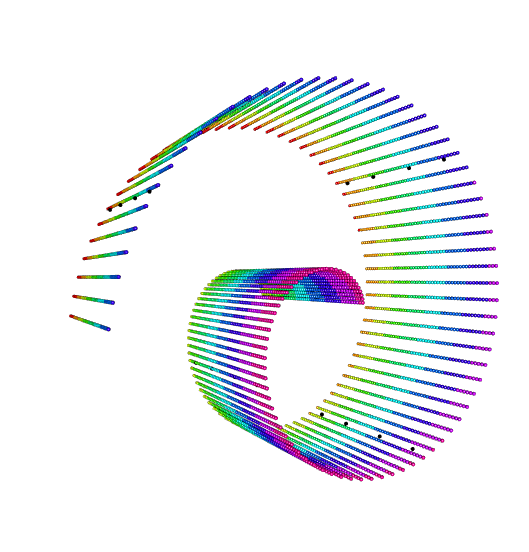}
\includegraphics[width=4.6cm]{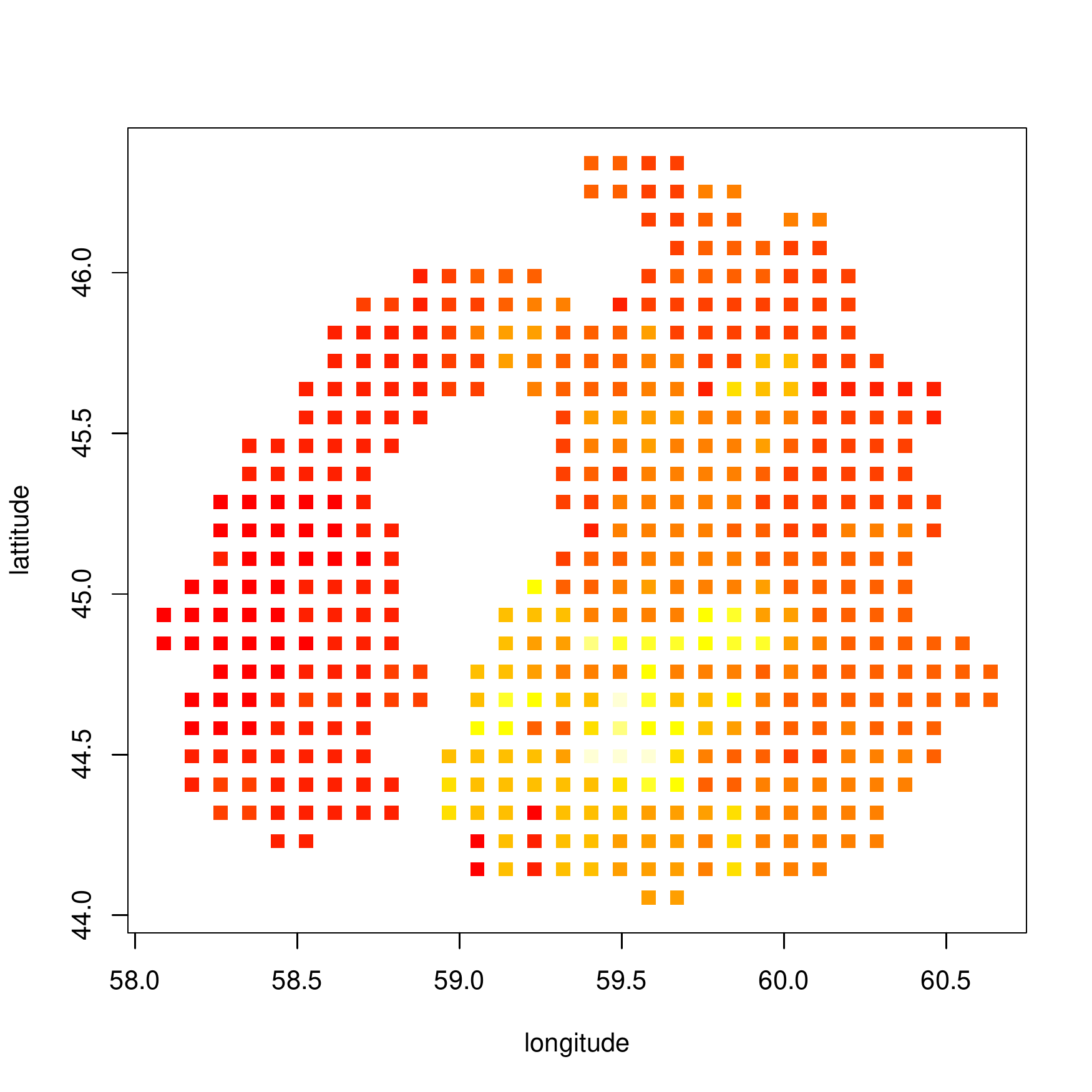}
    \caption{\label{fig-data1}
    %\footnotesize
    {Some illustrative examples.
     } } 
\end{figure}

In recent years it has become commonplace to collect data that are restricted to a complex constrained space.  For example, data may be collected in a spatial domain but restricted to a complex or intricately structured region corresponding to a geographic feature, such as a lake.  To illustrate, refer to the right panel of 
Figure \ref{fig-data1}, which plots satellite measurements on chlorophyll levels in the Aral sea \citep{wood}. 
In building a spatial map of chlorophyll levels in this sea, and in conducting corresponding inferences and prediction tasks, it is important to take into account the intrinsic geometry of the sea and its complex boundary.  Traditional smoothing or modelling methods that do not respect the intrinsic geometry of the space, and in particular the boundary constraints, may produce poor results.  For example, it is crucial to take into account the fact that pairs of locations having close Euclidean distance may be intrinsically far apart if separated by a land barrier.

Refer in particular to the locations near longitude 58.5 and 59 in the southern region of the map.  These locations have quite different chlorophyll levels due to the land barrier.  However, usual smoothing or modelling approaches that do not account for the boundary would naturally provide  close estimates of the chlorophyll level given their close spatial vicinity.   The goal of this article is to provide a general methodology that can accommodate not just complex spatial subregions of $\R^2$ (refer also to the U-shaped constraint in the left panel of Figure \ref{fig-data1}) but also complex subregions of higher-dimensional space ($\R^3$ and beyond) and usual manifold constraints, such as the Swiss roll in the middle panel of the figure.

To accommodate modelling on these broad and complex domains, we propose a novel class of {\em intrinsic} Gaussian processes (in-GPs).  in-GPs are designed to be useful in interpolation, regression and classification on manifolds, with a particular emphasis on complex or difficult regions arising as submanifolds.  in-GPs incorporate the intrinsic structure or geometry of the space, including the boundary features and interior conditions.  A major challenge in constructing GPs on manifolds is choosing a valid covariance kernel - this is a non-trivial problem and most of the focus has been on developing covariance kernels specific to a particular manifold (e.g,. \cite{guinness} consider low-dimensional spheres).
%and \cite{wood} some sub-manifolds of $\mathbb{R}^2$).
  Castillo et al (2014) instead proposed to use randomly rescaled solutions of the heat equation to define a valid covariance kernel for reasonably broad classes of compact manifolds.  They additionally provided lower and upper bounds on contraction rates of the resulting posterior measure.  Unfortunately, they do not provide a methodology for implementing their approach in practice, and their proposed heat kernels are computationally intractable.  

This article proposes a practical and general in-GP methodology, which uses heat kernels as covariance kernels.  This is made possible by the major novel contribution of the paper, which is to utilise connections between heat kernels and transition densities of Brownian motion on manifolds to obtain algorithms for approximating covariance kernels.  Specifically, the covariance kernels are approximated by first simulating a Brownian motion on the manifold or complex constrained space of interest, and then evaluating the transition density of the Brownian motion. 

Most current methods that can smooth noisy data over regions with a boundary can only be applied to spaces that are \emph{subsets of $\mathbb{R}^2$}; refer to \cite{wood} and \cite{tramsay}. \cite{ramsay13} extended  \cite{tramsay}'s smoothing spline method to model the brain surface arising as a subset of $\mathbb{R}^3$ by first discretising the surface. The main idea in this literature is to develop smoothing splines that respect the boundary or interior constraints.  Our in-GP approach is fundamentally different conceptually, while also having general applicability beyond two dimensional examples.  Although in-GPs will have an increasing computational cost as the dimensionality of the space increases, due to the need to simulate Brownian motion, there is no discretisation of the space unlike methods proposed in \cite{tramsay} and \cite{ramsay13}.   
 
Some other related work include  \cite{bruno} who extend kernel regression to a general Riemannian manifold.  \cite{david-regression} proposed to model a response and covariate on a manifold jointly using a Dirichlet process mixture model. The focus of our work on the other hand aims to generalise the powerful GP model to manifold-valued data. Although GPs have been extensively used in statistics and machine learning (see e.g., \cite{Rasmussen2004}), these models can not be directly generalised to model data on manifolds, such as irregular shape spaces, due to the difficulty of constructing valid covariance kernels. \cite{extrinsicGP} propose \emph{extrinsic covariance kernels} on general manifolds by first embedding the manifolds onto a higher-dimensional Euclidean space, and constructing a covariance kernel on the images after embedding.  However, such embeddings are not always available or easy to obtain for complex spaces.  

Section \ref{sec:2} discusses the construction of covariance kernels on manifolds and explores the connection between the heat kernel on a Riemannian manifold and the transition density of Brownian motion on the manifold.  This connection is utilised in developing practical algorithms for approximating the heat kernel. Section 3 focuses on inference under in-GPs using the approximated heat kernel of Section 2 
as the covariance kernel, including an extension to {\em sparse} in-GPs. Section 4 illustrates our in-GP methodology with various simulation and data examples. Section 5 contains a discussion.

\section{Intrinsic Gaussian process (in-GPs) on manifolds}
\label{sec:2}

\subsection{in-GPs with heat kernel as the covariance kernel}
  \label{GPmani}
  
  We propose to construct intrinsic Gaussian processes (in-GPs) on manifolds and complex constrained spaces using the heat kernel as the covariance kernel.  To be more specific, let $M$ be a $d$-dimensional complete and orientable Riemannian manifold, $\partial M$ its boundary,  $\Delta$  the Laplacian-Beltrami operator  on  $M$, and $\delta$ the Dirac delta `function'. 
  Heat kernels can be fully characterised as solutions to  the \emph{heat equation} with the \emph{Neumann boundary condition}:
\begin{align*}
\frac{\d}{\d t}K_{heat}(s_0,s,t)=\frac{1}{2}\Delta_s K_{heat}(s_0,s,t),\qquad
K_{heat}(s_0,s,0)=\delta(s_0,s), \ \ s\in M.
\end{align*}
  
 Alternatively, the  heat kernel $K_{heat}(x,y,t)\in C^{\infty}(M\times M\times \R^{+})$, the space of smooth functions on $M\times M\times \R^{+}$,  gives rise to an operator and satisfies
  \begin{align}
  (e^{\Delta t}f)(x)=\int_M K_{heat}(x,y,t)f(y)dy,
  \end{align}
  for any $f\in L^2(M)$.
  The heat kernel is symmetric with $K_{heat}(x,y, t)=K_{heat}(y,x,t)$, and is a positive semi-definite kernel on $M$ for any fixed $t$, and thus can serve as a valid covariance kernel for a Gaussian process on $M$. The \emph{Neumann boundary condition} can be expressed as no heat transfer across the boundary $\partial M$.
  
  If $M$ is a Euclidean space $\bb{R}^d$, the heat kernel has a closed form corresponding to a time-varying Gaussian function:
\begin{align*}
K_{heat}(\mb{x}_0,\mb{x},t)
=\frac{1}{(2\pi t)^{d/2}}\,
  \exp\left\{-\frac{||\mb{x}_0-\mb{x}||^2}{ 2t }\right\}, \; \mb{x}\in \mathbb R^d.
\end{align*}
In addition, the heat kernel of $\bb R^d$ can be seen as the scaled version of a radial basis function (RBF) kernel (or the popular squared exponential kernel) under different parametrisations: 
\begin{align*}
K_{RBF}(\mb{x}_0,\mb{x},l)
=\sigma_r^2\,
  \exp\left\{-\frac{||\mb{x}_0-\mb{x}||^2}{ 2l^2 }\right\}, \; \mb{x}\in \mathbb R^d.
\end{align*}

%{\color{red} But solving the heat equation on an arbitrary $\mathbb{M}$ is non-trivial, as discussed in section \ref{intro} we may not have a closed form solution for most cases. From now on, we denote BM for Brownian Motion and}

Letting $K_{heat}^t(x,y)=K_{heat}(x,y, t)$, our in-GP uses $K_{heat}^t(x,y)$ as the covariance kernel, where the  time parameter $t$ of $K_{heat}$ has a similar effect as that of the length-scale parameter $l$ of $K_{RBF}$, controlling the rate of decay of the covariance.  By varying the time parameter, one can vary the bumpiness of the realisations of the in-GP over $M$. 

We use in-GPs to develop nonparametric regression and spatial process models on complex constrained domains $M$.  Let $\mathcal D = \{ (s_i,y_i), i =1,\ldots,n \}$ be the data, with  $n$ the number of observations, $s_i \in M$ the predictor or location value of observation $i$ and $y_i$ a corresponding response variable.  We would like to do inferences on how the output $y$ varies with the input $s$, including predicting $y$ values at new locations $s_*$ not represented in the training dataset.  Assuming Gaussian noise and a simple measurement structure, we let 
\begin{align}
y_i = f(s_i) + \epsilon_i, \  \  \    \epsilon_i \sim \mathcal N(0, \sigma_{noise}^2),  \  \  \ s_i\in M,
\end{align}
 where $\sigma_{noise}^2$ is the variance of the noise.  This model can be easily modified to include parametric adjustment for covariates $x_i$, and to accommodate non-Gaussian measurements (e.g., having exponential family distributions).  However, we focus on the simple Gaussian case without covariates for simplicity in exposition.
 
Under an in-GP prior for the unknown function $f:M \to \Re$, we have 
\begin{align}
p(\text{{\bf f}} | s_1,s_2,...,s_n) = \mathcal N( \bm{0},\Sigma),
\end{align}
where {\bf f} is a vector containing the realisations of $f(\cdot)$ at the sample points $s_1,\ldots,s_n$, $f_i = f(s_i)$, and 
$\Sigma$ is the covariance matrix in these realisations induced by the in-GP covariance kernel. In particular, the entries of $\Sigma$ are obtained by evaluating the covariance kernel at each pair of locations, that is,
 \begin{align}
 \label{sk:heat}
 \Sigma_{ij} = \sigma_h^2 K_{heat}^t(s_i,s_j).
 \end{align}
Following standard practice for GPs, this prior distribution is updated with information in the response data to obtain a posterior distribution.  Explicit expressions for the resulting predictive distribution are provided 
 in Section \ref{BMinGP}.
\begin{remark}
We added an additional hyperparameter $\sigma_h^2$ by rescaling the heat kernel for extra flexibility.  The parameter $\sigma_h^2$ plays a similar role as that of the magnitude parameter of a standard squared exponential kernel in the Euclidean space. As mentioned above,  the parameter $t$ is analogous to the length-scale parameter in a squared exponential or RBF kernel.  
\end{remark}

\subsection{Numerical approximation of the heat kernel: exploiting connections with the transition density of Brownian motion} \label{BMinGP}

One of the key challenges for inference using in-GPs with the construction in Section \ref{GPmani} is that \emph{closed form expressions for $K_{heat}^t$ do not exist for general Riemannian manifolds}.  
Explicit solutions are available only for very special manifolds such as the Euclidean space or spheres. Therefore, for most cases, one can not explicitly evaluate $K_{heat}^t$ or the corresponding covariance matrices.  To overcome this challenge and  bypass the need to solve the heat equation directly, we utilise the fact that heat kernels can be interpreted as \emph{transition density functions of Brownian motion (BM) in $M$. } Our recipe is to simulate Brownian motion on $M$, numerically evaluate the transition density of the Brownian motion, and then use the evaluation to approximate the kernel $K_{heat}^t(s_i,s_j)$ for any pair $(s_i, s_j)$. 

To explain explicitly the equivalence between the heat kernel and the transition density of the BM, let $S(t)$ denote a BM on $M$ started from $s_0$ at time $t=0$.  The probability of $S(t) \in A \subset M$, for  any Borel set $A$, is  given by 
\begin{align}
\label{BMprob}
\bb{P}\big[S(t)\in A\,|\,S(0)=s_0\big] = \int_A K_{heat}^t(s_0,s)ds,
\end{align}
where the integral is defined with respect to the volume form of $M$.  % I guess this means that we are implicitly assuming that the manifold M is orientable?  does this rule out manifolds we are likely to encounter in statistics and should we state such assumptions up front?  What if $M$ is some sort of complicated subset of Euclidean space that is not a Riemannian manifold?
In this context, the Neumann boundary condition simply means that, whenever $S$ hits $\d M$, it keeps going within $M$.

We approximate the heat kernel via approximating the integral in equation \eqref{BMprob} by simulating BM sample paths and numerically evaluating the transition probability. Considering the BM $\{S(t):t>0\}$ on $M$ with the starting point $S(0)=s_0$, we simulate $N$ sample paths. For any $t>0$ and $s\in M$, the probability of $S(t)$ in a small neighbourhood $A$ of $s$  can be estimated by counting how many BM sample paths reach $A$ at time $t$.  Note that the BM diffusion time $t$  works as the smoothing parameter. If $t$ is large, the BM has higher probability to reach the neighbourhood of the target point and leads to higher covariance and vice versa.

\begin{figure}[htb]
    \centering{
\includegraphics[width=0.5\textwidth,height=0.4\textwidth]{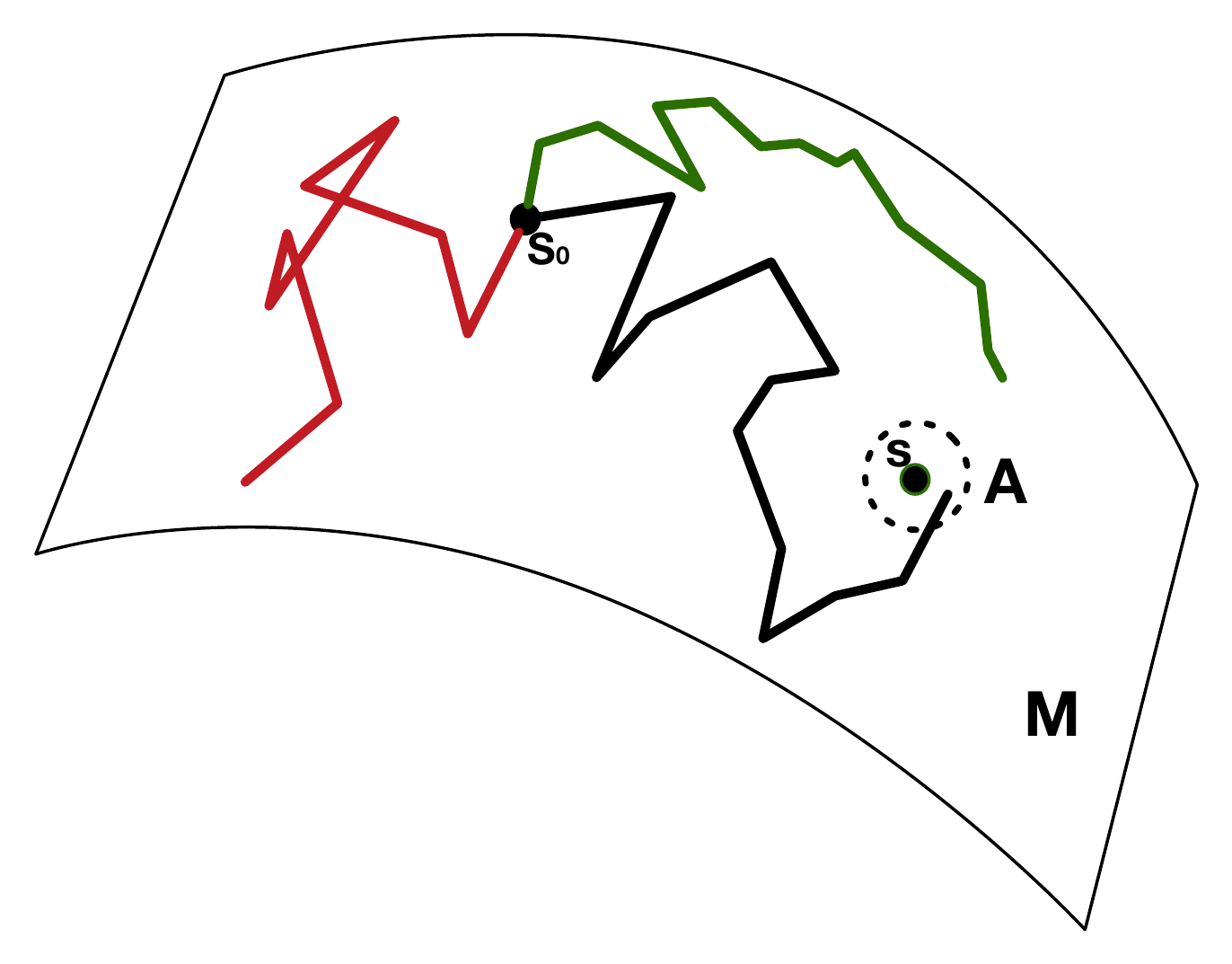}
    \caption{ \label{fig:bmM}
     BM on a manifold $M$.  $s_0$ is the starting point of BM sample paths. The solid lines represent three independent BM sample paths from time $0$ to $t$.  The dashed circle represents a set $A$, which is a neighbourhood of a point $s$ on $M$.  In this example, only the black sample path reaches $A$ at time $t$ and the estimate of the transition probability $p(S(t)\in A | S(0)=s_0)$ is $\frac{1}{3}$.
     } }
\end{figure}

The \emph{transition probability} is approximated as 
\begin{align}
\bb{P}\big[S(t)\in A\,|\,S(0)=s_0\big]  \approx \,\bb{P}\big[\|(S(t)-s\|<w\big] =  \frac{k}{N},
\end{align}
where $\|\cdot\|$ is the Euclidean distance, $w$ is the radius determining the Euclidean ball size, $N$ is the number of the simulated BM sample paths and $k$ is the number of BM sample paths which reach $A$ at time $t$. 
%When $w$ is small, $\rho(S(t),s)$ {\color{red}can be accurately approximated as the Euclidean distance between $S(t)$ and $s$}. 
An illustrative diagram is shown in Figure \ref{fig:bmM}. The \emph{transition density} of $S(t)$ at $s$ is approximated as 
\begin{align}
\label{k:heat}
K_{heat}^t(s_0,s) \approx \hat{K}^t  =\frac{1}{V(w)}\,\bb{P}\big[\|S(t)-s\|<w\big] = \frac{1}{V(w)}\cdot \frac{k}{N},
\end{align}
where  $V(w)$ is the volume of $A$, which is parameterised with $w$, and $\hat{K}$ is the estimated transition density. 
%The distance on manifold between $s_0$ and $s$ is modeled implicitly in $K_{heat}(s_0,s,t)$ in terms of the BM transition probability from $s_0$ to $s$. 
 The error (numerical error and Monte Carlo error) of this estimator of the heat kernel is discussed in section \ref{estierror}. Discussions on how to simulate BM sample paths on manifolds are deferred to section \ref{trandens}. 

 The in-GP can be constructed using the approximation in \eqref{k:heat}. The covariance matrix of the training data  $\Sigma_{ \text{\bf ff} }$ can be explicitly obtained as follows:
 for the $i_{th}$ row of $\Sigma_{ \text{\bf ff} }$, $N$ BM sample paths are simulated, with the starting point the $i_{th}$ data point indexed by the corresponding row. For each element of the $i_{th}$ row, $\Sigma_{ \text{\bf ff} }$, $\hat{K}^t(s_i,s_j)$ is then estimated using \eqref{k:heat}. 
  Algorithm \ref{alg:bm} below provides details on how to generate $\Sigma_{\text{\bf ff}}$.
 %The algorithm of generating $\Sigma_{\text{\bf ff}}$ is shown in algorithm \ref{alg:bm}. 

 \begin{algorithm}[!h]
\caption{ { Simulating Brownian motion sample paths for estimating $\Sigma$ } }
\label{alg:bm}
\begin{algorithmic}
\State 1.1 Generate Brownian motion sample paths
\For{$i = 1,\ldots,N_d$}    \Comment{ {\small $N_d$ is the size of data points}  }
\For{$j = 1,\ldots,N_{bm}$}    \Comment{ {\small$N_{bm}$ is No. of sample paths} }
\For{$ l = 1,\ldots,T$}    \Comment{ {\small T steps Brownian motion, $T*\Delta t$  $\rightarrow$ max diffusion time} }
%\State                                \Comment{$T*\Delta t$  $\rightarrow$ max diffusion time}
\State {\bf do}
\State $q\left(x_{i,j}(l)|x_{i,j}(l-1)\right) \leftarrow \mathbb N \left( x_{i,j}(l) |  \mu \left( x_{i,j}\left(l-1\right),\Delta t \right)  , \Delta t g^{-1} \right )$ \Comment{ {\small use eqn \ref{disProp} } } 
\State {\bf While} $x_{i,j}(l) \notin \partial M$  \Comment{ {\small keep proposing x until locating within boundary } } 
\EndFor
\EndFor
\EndFor
\Return $\bm x$
\State 1.2 Given a discrete choice of the diffusion time $t\in \{ \Delta t, 2*\Delta t,\cdot \cdot \cdot, T*\Delta t \}$, the covariance matrix $\Sigma^{t}$ is estimated based on the BM simulation from 1.1.
\For{$i = 1,\ldots,N_d$} 
\For{$j = 1,\ldots,N_d$} 
\State k = which( $\|x(t) - s_j\| < w$ )  \Comment{ {\small counting how many BM paths reach $A_{s_j}$}  }
\State $K_{heat}^t(s_i,s_j)= \frac{ k}{N_{BM}*V(w)}$  \Comment{ {\small use eqn \ref{k:heat} } }
\State $\Sigma_{ij}^t = \sigma_h^2 K_{heat}^t(s_i,s_j) $
\EndFor
\EndFor
\Return $\Sigma^t$
\end{algorithmic}
\end{algorithm}
 
Optimisation of the kernel hyper parameters is discussed in section \ref{optimT}.

Given in-GPs as the prior, one can then update with the likelihood to obtain the posterior distribution for inference. Let  ${\bf f}_*$ be a vector of values of $f(\cdot)$ at some test points not represented in the training sample. The joint distribution of ${\bf f} $ and ${\bf f}_*$ are Gaussian:

\begin{align}
p( {\text {\bf f}}, {\text {\bf f}}_*) = \mathcal{N} \left ( 0,
 \left [ \begin{array}{cc}  
 \Sigma_{{\text {\bf f}} {\text {\bf f}}} &\Sigma_{ {\text {\bf f}}  {\text {\bf f}}_*} \\[0.3em]
 \Sigma_{ {\text {\bf f}}_* {\text {\bf f}}} &\Sigma_{ {\text {\bf f}}_* {\text {\bf f}}_*} \\[0.3em]
                \end{array} \right ]
                \right ),
\end{align}
where $\Sigma_{{\text {\bf f}}_*{\text {\bf f}}}$ is the covariance matrix for training data points and test points. Each entry of the covariance matrix of the joint distribution can be calculated using equation \eqref{sk:heat}:
 \begin{align}
 \label{sk:heat}
 \Sigma_{_{ij} }= \sigma_h^2 \hat{K}^t(s_i,s_j).
 \end{align}
%Both $\Sigma_{{\text {\bf f}} {\text {\bf f}}}$ and $\Sigma_{ {\text {\bf f}}  {\text {\bf f}}_*}$ can be estimated from the BM sample paths with the starting points of which are the training set points.
For the same row of $\Sigma_{{\text {\bf f}} {\text {\bf f}}}$ and $\Sigma_{ {\text {\bf f}}  {\text {\bf f}}_*} $, all elements can be estimated from the same patch of BM simulations which share the same starting points. We do not need additional BM simulations to estimate $\Sigma_{ {\text {\bf f}}  {\text {\bf f}}_*} $.
The predictive distribution is derived by marginalising out {\bf f}:

\begin{align}
p( {\text {\bf f}}_*| \bm y) = \int p( {\text {\bf f}}_*{\text {\bf f}}| \bm y) d{\text {\bf f}} = \mathcal N \left ( \Sigma_{ {\text {\bf f}}_*{\text {\bf f}}}  \left(\Sigma_{{\text {\bf f}}{\text {\bf f}}}  + \sigma_{noise}^2 I \right)^{-1}\bm y , \ \  \Sigma_{{\text {\bf f}}_*{\text {\bf f}}_*}-\left(\Sigma_{{\text {\bf f}}{\text {\bf f}}} +\sigma_{noise}^2 I \right)^{-1} \Sigma_{{\text {\bf f}}{\text {\bf f}}_*} \right).
\end{align}
If we are only interested in the predictive mean, only $\Sigma_{{\text {\bf f}}_*{\text {\bf f}}}$  and $\Sigma_{{\text {\bf f}}{\text {\bf f}}}$ need to be estimated.  
The predictive variance of test points requires computing the covariance matrix $\Sigma_{{\text {\bf f}}_*{\text {\bf f}}_*}^t$. This requires extra BM simulations whose starting points are the test points. This could be computationally heavy if the number of test points is big. The sparse in-GP  is introduced in the next section to handle this problem.

\subsection{ Sparse in-GP on manifolds to reduce computation cost} \label{sec:sparseGP}

The construction of  in-GPs proposed in section \ref{BMinGP} requires simulating BM sample paths at  each data point. Although the BM simulations are  embarrassingly parallelizable, the computational cost can be high when the sample size is large. In addition, Gaussian processes face the well-known problem of  high-computational complexity $O(n^3)$ due to the inversion of the covariance matrix. In this section, we propose to combine in-GPs with sparse Gaussian process approximations proposed by \cite{quin07} to alleviate the complexity problems. We call the resulting construction \emph{sparse in-GP}.  By employing sparse in-GP, Brownian motion paths only need to be simulated \emph{starting  at the induced points}  instead of every data point.

The GP prior can be augmented with an additional set of $m$ inducing points on $M$ denoted as $\bm{z}=[z_1,...,z_m]$, $z_i\in M$ and we have $m$ random variables $\bm{u}=[f(z_1),...,f(z_m)]$. The marginal prior distribution $p({\bf f}_*,{\bf f})$ remains unchanged after the model being rewritten in terms of the prior distribution $p(\bm{u})$ and the conditional distribution $p({\bf f}_*, {\bf f} |\bm{u})$:
\begin{align}
p({\bf f}_{*},{\bf f})& = \int p({\bf f}_*,{\bf f},\bm{u})d\bm{u} =\int p({\bf f}_{*},{\bf f}|\bm{u})p(\bm{u})d\bm{u}, \\
p(\bm{u}) &= \mathcal N (0, \Sigma_{ {\bf uu}}), \label{eqn:prior_u}
\end{align}
where the distribution of $u$ is a multivariate Gaussian with mean zero and covariance matrix $\Sigma_{\bf uu}$. The above augmentation does not reduce the computational complexity. For efficient inference, we adopt  the Deterministic Inducing Conditional approximation by \cite{quin07}, where ${\bf f}_*$ and ${\bf f}$ are assumed to be conditionally independent given $\bm{u}$ and the relations between any ${\bf f}$ and $\bm{u}$ are assumed to be deterministic:
\begin{align}
\label{app:int}
p({\bf f}_*,{\bf f}) \approx q({\bf f}_*,{\bf f}) &=\int q({\bf f}_*|{\bf u})q({\bf f}|\bm{u})p(\bm{u})d\bm{u}, \\
q({\bf f}|\bm{u}) &= \mathbf{N}(\bm{\mu_f},0), \ \  \bm{\mu_f} = \Sigma_{{\bf fu}} \Sigma_{{\bf uu}}^{-1}\bm{u}, \\
q({\bf f}_*|\bm{u})& = \mathbf{N}(\bm{\mu_*},0), \ \  \bm{ \mu}_* = \Sigma_{{\bf f_*u}} \Sigma_{{\bf uu}}^{-1}\bm{u}.
\end{align}
The resulting sparse in-GP prior by taking the Deterministic Inducing Conditional  approximation is also written as 
\begin{align*}
  q({\bf f,f_*})= \mathbf{N} \left( 0, \left[\begin{array}{cc}  
Q_{\bf ff} & Q_{\bf ff_*} \\[0.3em]
Q_{\bf f_*f} & Q_{\bf f_*f_*} \\[0.3em]
           \end{array} \right] \right) =\mathbf{N} \left( 0, \left[\begin{array}{cc}  
\Sigma_{\bf fu} \Sigma_{\bf uu}^{-1} \Sigma_{\bf uf} & \Sigma_{\bf fu} \Sigma_{\bf uu}^{-1} \Sigma_{\bf uf_*} \\[0.3em]
\Sigma_{\bf f_*u} \Sigma_{\bf uu}^{-1} \Sigma_{\bf uf} & \Sigma_{\bf f_*u} \Sigma_{\bf uu}^{-1} \Sigma_{\bf uf_*} \\[0.3em]
           \end{array} \right] \right)
\end{align*}

\noindent where $Q$ is defined as  $Q_{\bf a,b} = \Sigma_{\bf a, u}\Sigma_{\bf u,u}^{-1}\Sigma_{\bf u,b}$. Using algorithm \ref{alg:bm}, $\Sigma_{ {\bf uu}}$, $\Sigma_{{\bf uf}}$ and $\Sigma_{{\bf uf_*}}$ are all obtained by estimating the transition density of BM simulation paths with inducing points as the starting points. 

With Deterministic Inducing Conditional  approximation, we only need to simulate the BM sample paths starting from the inducing points. The total number of BM simulations is reduced from $n\times N_{bm}$ to $m\times N_{bm}$, where $m$ is the number of inducing points, $n$ is the number of data points and $N_{bm}$ is the number of Brownian motion sample paths given a single starting point. The complexity of inverting the covariance matrix is also decreased from $O(n^3)$ to $O(n\times m^2)$.

With the above approximation, the marginal distribution of the corresponding GP with a Gaussian likelihood is written as:
\begin{equation}
p(\bm{y}|{\bf f}) \approx q(\bm{y}|\bm{u}) = \prod_{i=1}^{n} \mathbf{N} \left( y_i| \Sigma_{f_i \bf u} \Sigma_{\bf uu}^{-1} {\bf u} , \sigma_{noise}^2 {\bf I} \right). 
\end{equation}

The inducing points in the above marginal likelihood can be further marginalised out by substituting the definition of its prior distribution (\ref{eqn:prior_u}):
\begin{equation} \label{sparseLik}
p(\bm{y}|\bm{s}_{induce}) = \int q(\bm{y}|\bm{u}) p(\bm{u}|\bm{s}_{induce})d\bm{u} = \mathbf{N} \left( 0, \Sigma_{\bf fu} \Sigma_{\bf uu}^{-1} \Sigma_{\bf uf}+\sigma_{noise}^2 {\bf I} \right).
\end{equation}

With the above model, we can also obtain the predictive distribution as
\begin{align}
q( {\bf f}_*| \bf{y} ) &= \mathbf{N} \left( Q_{\bf f_*f}  \left( Q_{\bf ff}+ \sigma^2 \bf{I} \right)^{-1} {\bf y} , Q_{\bf f_*f_*}-Q_{\bf f_*f}(Q_{\bf ff}+\sigma^2 {\bf I})^{-1} Q_{\bf ff_*} \right).
%q( \mathbf{f_{*} }| \bf{y} ) &= \mathbf{N} \left( Q_{ {\bf f_*f}}  \left( Q_{\mathbf{ff}}+ \sigma^2 \bf{I} \right)^{-1} {\bf y} , Q_{ {\bf f_*f_*}}-Q_{ {\bf f_*f}}(Q_{ {\bf ff} }+\sigma^2 {\bf I})^{-1} Q_{ {\bf ff_*} } \right).
%\mathbf{Q_{ff}} &= \Sigma_{fu} \Sigma_{uu}^{-1} \Sigma_{uf} \\
%\mathbf{Q_{f_*f}} &= \Sigma_{f_{*}u} \Sigma_{uu}^{-1} \Sigma_{uf}
\end{align}

Apart from the Deterministic Inducing Conditional (DIC), there is a huge literature on reducing the matrix inversion bottleneck in GP computation \citep{SchwaighoferTresp2002, QuioneroCandelaRasmussen2005, SnelsonGhahramani2006,Titsias2009}. Recent approaches, such as \cite{katzfuss2017}, can achieve linear time computation complexity under certain conditions.  However, such approaches require an analytical form of covariance kernel; to apply these methods we would need to simulate BM paths at the training and prediction points.  For this reason, we use DIC due its avoidance of the need to estimate the diagonal elements of the covariance matrix.

%{\color{red}   mention other approximation?? 
%
%appendix ??
%low rank approximation of the full covariance matrix $\Sigma_{ff}$ and  approximate the likelihood with the projection of f on u \citep{quin07}, $f=\Sigma_{fu} \Sigma_{uu}^{-1}u$
%\begin{align}
%\Sigma_{ff} \approx \Sigma_{fu} \Sigma_{uu}^{-1} \Sigma_{uf}
%\end{align}
%
%how to derive predictive distribution:
%\begin{align*}
%p(y|f) = \mathbf{N}(f,\sigma^2 \mathbf{I}) \ \ \  p(u) = \mathbf{N}(0,k_{uu}) \\
%?? q(f^*|y) = \int q(f^*|f,y)df
%\end{align*}
%
%\begin{align}
%q(f^{*}| \bf{y} ) &= \mathbf{N} \left( \mathbf{Q_{*f}} \left( \mathbf{Q_{ff}}+ \sigma^2 \bf{I} \right)^{-1}\bf{y} , Q_{**}-Q_{*f}(Q_{ff}+\sigma^2 \bf{I})^{-1} Q_{f*} \right)
%\end{align}
%}

\subsection{Monte Carlo error and  numerical error for the approximation of heat kernel } 
\label{estierror}

In this subsection, we discuss the error of our  heat kernel estimator  as defined in equation \eqref{k:heat}. 
We also consider numerical experiments in the special case of $\mathbb R$ in which case the true heat kernel is known. 

Consider a Brownian motion $\{S(t):t>0\}$ on a Riemannian manifold $M$ with $S(0)=s_0$.
Fix some $t>0$ and $s\in M$. The probability density of $S(t)$ at $s$ is $K_{heat}^t(s_0,s)$. The true BM transition probability evaluated at a set $A$  is given by $p(A)=\bb{P}\big[S(t)\in A\,|\,S(0)=s_0\big] =\int_A K_{heat}^t(s_0,s) ds.$ The error of our estimator $\hat K^t$ consists of two parts.

{\noindent {\bf Part I: Numerical error}. Choose local coordinates $(r_1,\ldots,r_d)$ near $s$ with $r_1(s)=\ldots=r_d(s)=0$ (for convenience of illustration) and a window size $w$.
The heat kernel $K_{heat}^t$ can then be approximated  by
\begin{align*}
{K^t}' =\frac{1}{V(w)}\,\bb{P}\Big[|r_i(S(t))|<w\textrm{ for }i=1,\ldots,d\Big],
\end{align*}
where $V(w)$ denotes the volume of the region defined by $\{|r_i|<w, i=1,\ldots,d\}.$
By Taylor expansion around $s$, we have
\begin{align}
{K^t}'=K_{heat}^t+O(w^2).
\end{align}
Therefore, the approximation error increases (quadratically) with $w$, i.e.,  the order of magnitude of ${K^t}'-K_{heat}^t$ is $O(w^2)$.}

If $M=R^d$, one can explicitly derive the error. Assume the starting point of BM $s_0$ is the origin for simplicity, the heat kernel $K_{heat}^t$ on $R^d$ can be approximated as
\begin{align}
\label{eqn:2ndapprox}
{K^t}'  \nonumber&=\frac{1}{V(w)}\,\bb{P}\big[\|S(t)-s\|<w\big] = \frac{1}{V(w)}\,  \int_A K_{heat}^t(s_0,s)ds,\\
& = \frac{1}{(2w)^d}\,  \int_{s_1-w}^{s_1+w} ... \int_{s_d-w}^{s_d+w} exp\left( \frac{-\sum_{i=1}^d x_i^2}{2t}  \right)dx_d...dx_1.
\end{align}
%\begin{align}
%\label{eqn:ke}
%K' = \frac{1}{V(w)}\,  \int_A K ds = \frac{1}{(2w)^d}\,  \int_{s_1-w}^{s_1+w} ... \int_{s_d-w}^{s_d+w} exp\left( \frac{-\sum_{i=1}^d x_i^2}{2t}  \right)dx_d...dx_1 ,
%\end{align}
Taylor expansion of equation \eqref{eqn:2ndapprox}  yields
\begin{align}
\label{eqn:tay}
{K^t}' - K_{heat}^t =  \frac{ \sum_{i=1}^d{s_i^2} -  d \cdot t  }{6t} \cdot\frac{w^2}{t}
  +O\left(\frac{w^4}{t^2}\right).
\end{align}
%In particular, the approximation error increases (quadratically) with $w$.
 Assuming $w$ is small compare to $\sqrt{t}$,  the order of magnitude of this error is $O(w^2)$.

\begin{remark}
For convenience in computing the integral in equation \eqref{eqn:2ndapprox}, a hypercube is used instead of the Euclidean ball. The order of the magnitude of the error remains the same.
\end{remark}

{\noindent {\bf Part II: Monte Carlo error.}  Given  $N_{BM}$ number of BM sample paths, ${K^t}'$ is approximated by $\hat{K}^t$:
\begin{align}
\label{eqn:1stapprox}
\hat{K}^t = \frac{1}{V(w)}\cdot\frac{k}{N_{BM}},\quad
\t{where }k\sim\t{Bin}(N_{BM},V(w) {K^t}' ).
\end{align}
Recall that $k$ is the number of sample paths within $\|S(t)-s\|<w$ and has binomial distribution with  $N_{BM}$ trials and probability of success $V(w) {K^t}'$. Here $\frac{k}{N_{BM}}$ is the estimate of the  transition probability of BM.
The standard error of  $\hat{K}^t$ is
\begin{align}
\label{eqn:bin}
sd\left( \hat{K}^t \right) &\nonumber= \frac{1}{N_{BM}V(w)}\cdot\sqrt{N_{BM}\cdot V(w) {K^t}' (1-V(w){K^t}')} \\
&\leq \sqrt{ \frac{ {K^t}' }{ N_{BM}V(w)}  } =O(w^{-d/2}),
\end{align}
which decreases with $w$ and $N_{BM}$.}

 The optimal order of magnitude of $w_{opt}$ can be calculated by minimising the sum of the two errors described above. Specifically, for an arbitrary $M$, one has
 \begin{align}\label{esterror}
\hat{K}^t - K_{heat}^t = O(w^2) + O(w^{-d/2}).
\end{align}
In particular if $M$ is $R^d$, an explicit expression of the error is available: 
\begin{align}
\text{estimation error} = \sqrt{ \frac{ \hat{K}^t }{ N\cdot (2w)^d} } + K_{heat}^t \frac{ \sum_{i=1}^d{s_i^2} -  d \cdot t  }{6t} \cdot\frac{w^2}{t}.
\end{align}

Given a pre-specified error level, the order of the minimum number of BM simulations $N$ required can  be derived. Refer to Appendix A for the example of estimating the heat kernel in a one-dimensional Euclidean space.

\begin{figure}[htb]
    \centering
\includegraphics[width=0.5\textwidth,height=0.4\textwidth]{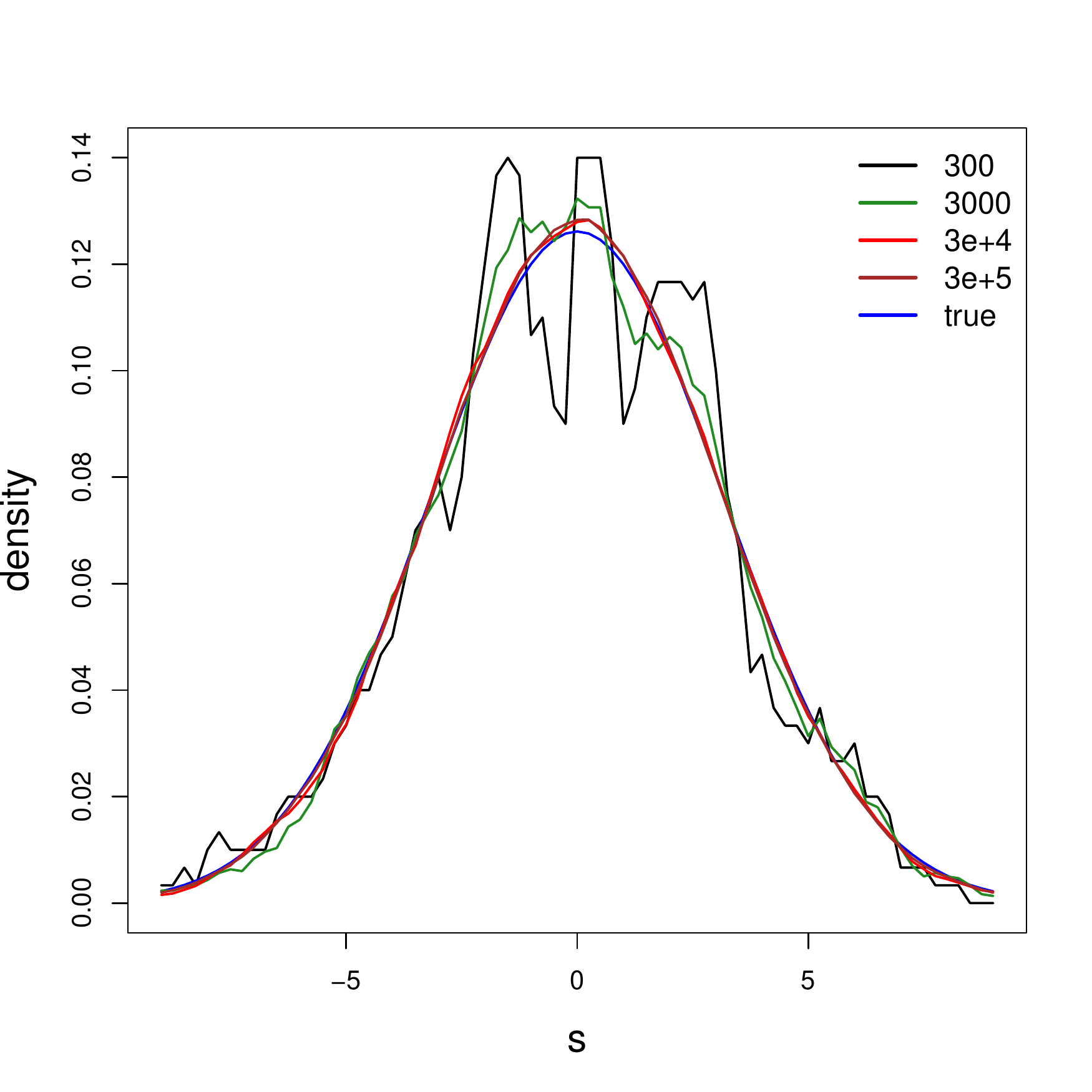}
    \caption{ \label{fig:bmTR}
    %\footnotesize
     Comparison of estimates of BM transition density and the heat kernel in $\mathbb R$. The blue line represents the true heat kernel. Colored lines represent estimates of the BM transition density given different number of BM simulations ranging from 300 to $3\times 10^5$.
     }
\end{figure}

Numerical accuracy of estimates for the special case of $\R$ are shown in Table \ref{tb:realL} and Figure \ref{fig:bmTR}. The true heat kernel $K_{heat}^t(0,s)$ is calculated using equation \eqref{eqn:trueR} at seventy equally spaced $s \in (-9,9)$. The diffusion time is fixed as 10. The transition probability of BM from the origin to the grid point $s$ is estimated by counting how many BM paths reach the neighbourhood of $s$ ($[s-w,s+w]$) at time $t$. The transition density of BM at each grid point is then evaluated using equation \eqref{k:heat}. Using equation \eqref{eqn:wop} the order of magnitude of $w_{opt}$ is derived as $10^{-1}$ for all grid points. We fix the radius $w$ as 0.5 in equation \eqref{k:heat}.

\begin{table}
\label{tb:realL}
\caption{Comparison of estimates of BM transition density  and the heat kernel in $\mathbb R$.  The table shows the  median absolute error and median relative error between the true heat kernel $K_{heat}^t$ and the numerical estimate of BM transition density. Values in brackets show the Median absolute deviation.}
\centering{
\fbox{
\begin{tabular}{*{4}{c}}
\em No. Sample paths  $N_{BM}$ &\em median absolute error&\em median relative error \\
\hline
3e+2  & 8.4e-3(8.9e-3)  & 24.6\%(25.6e-1)\\
3e+3  & 2.8e-3 (2.9e-3)  &6.4\% (5.5e-2)  \\
3e+4  & 7.2e-4 (6.8e-4) &1.6\% (1.9e-2)   \\
3e+5 & 4.7e-4 (3.8e-4) &1.3\% (1.1e-2)  \\
\end{tabular}}  }
\end{table}
The number of BM simulation sample paths $N_{BM}$  are selected from three hundred to three hundred thousand with increasing  order of magnitude. The median of relative error decreases as $N_{BM}$ increases and stabilises after thirty thousand. A similar pattern is observed for the median absolute error.  Derivations for the transition density estimate of heat kernel in $\mathbb{R}^2$ is shown in the Appendix.

\subsection{Simulating Brownian motion on manifolds}
\label {trandens}

\begin{figure}[htb]
    \centering
\includegraphics[width=0.7\textwidth,height=0.35\textwidth]{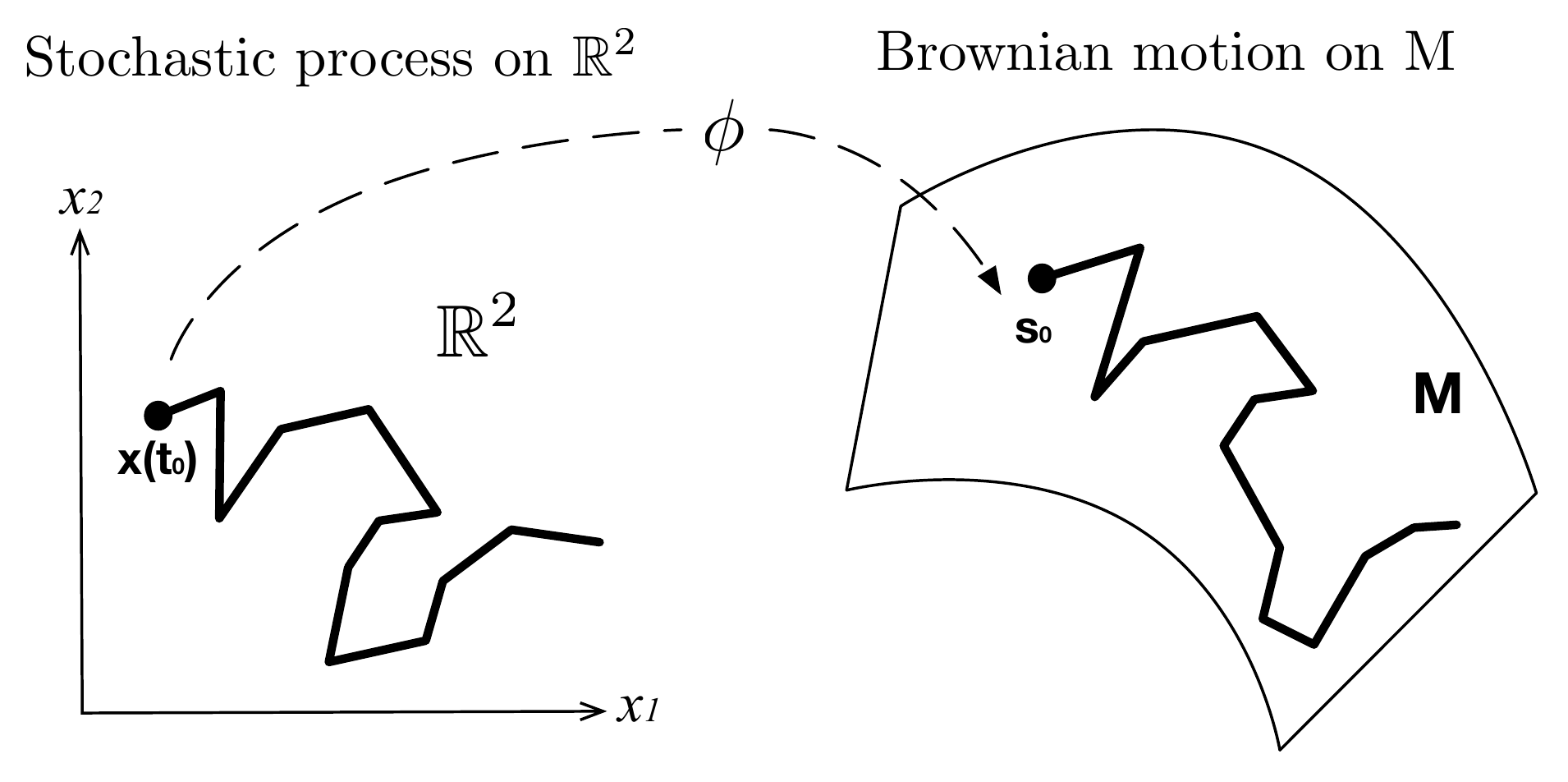}
    \caption{ \label{fig:metric}
    %\footnotesize
    { BM on $M$ and its equivalent stochastic process in local coordinate system in $\mathbb{R}^2$.  $\phi: \mathbb{R}^2 \rightarrow M$ is a local parametrisation of $M$.  }}
\end{figure}

In order to estimate the transition density of Brownian Motion (BM) on $M$, we first need to simulate BM sample paths on $M$.  Let  $\phi : \mathbb{R}^d \rightarrow M $ be a local parameterisation of $M$ around $s_0 \in M$ which is smooth and injective. A demonstration of $\phi$ is depicted in Figure \ref{fig:metric}.  Let $\mb{x}(t_0) \in R^2$ be such that $\phi \left( \mb{x} \left(t_0\right) \right)=s_0$. The Riemannian manifold $M$ is equipped with a metric tensor $g$ by defining an inner product on the tangent space:
\begin{align}
\label{metric}
g_{ij}( \mb{x})= \frac{ \partial \phi}{ \partial x_i} (\mb{x}) \cdot \frac{ \partial \phi}{ \partial x_j} (\mb{x}).
\end{align}

 As in Figure \ref{fig:metric}, simulating a sample path of BM on $M$ with starting point  $s_0$ is equivalent to simulating a stochastic process in $\mathbb{R}^2$ with starting point $\mb{x}(t_0)$.
The BM on a Riemannian manifold in a local coordinate system is given as \citep{hsu1988,hsu2008}
\begin{align}
\label{eqnBM}
dx_i(t) = G^{-1/2} \sum^{d}_{j=1}\frac{\partial}{\partial x_j} \left(  g_{ij}^{-1} G^{1/2} \right) dt + \left( g^{-1/2} dB(t)\right)_i
\end{align}
where $g$ is the metric tensor of $M$, $G$ is the determinant of $g$ and $B(t)$ represents an independent BM in the Euclidean space.
If $M=\R^d$, $g$ become an identity matrix and $x_i(t)$ is the standard BM in $\R^d$. The first term of equation (\ref{eqnBM}) is related to the local curvature of $M$ which becomes a constant if the curvature is a constant. The second term relates to the position specific alignment of the BM by transforming the standard BM $B(t)$ in $\mathbb{R}^d$ based on the metric tensor $g$.

For simulating BM sample paths, the discrete form of equation (\ref{eqnBM}) is first derived in equation \eqref{disBM}. Specifically, the Euler Maruyama method is used \citep{kloeden1992,lamberton2007} which yields:
\begin{align}
\label{disBM}
x_i(t) &= x_i(t-1) + \sum^{d}_{j=1} \left(  -g^{-1} \frac{\partial g}{\partial x_j} g^{-1} \right)_{ij} \Delta t + \frac{1}{2} \sum_{j=1}^d(g^{-1})_{ij}tr(g^{-1}\frac{\partial g}{\partial x_j}) \Delta t + \left( g^{-1/2} dB(t)\right)_i \nonumber \\
         &= \mu( x_i(t-1),\Delta t)_i + \left( \sqrt{\Delta t} g^{-1/2} z^d\right)_i,
\end{align}
 where $\Delta t$ is the diffusion time of each step of the BM simulation and $z^d$ in the second line of equation \eqref{disBM} represents a $d$-dimensional normal distributed random variable. The discrete form of the above stochastic differential equation defines the proposal mechanism of the BM with density
\begin{align}
\label{disProp}
q\left(x(t)|x(t-1)\right) = \mathbb N \left( x(t) |  \mu \left( x\left(t-1\right),\Delta t \right)  , \Delta t g^{-1} \right ).
\end{align}
This  proposal  makes BM move according to the metric tensor.

If the manifold $M$ has boundary $\partial M$, we apply the Neumann boundary condition as in section \ref{BMinGP}.
 It implies the simulated sample paths only exist within the boundary. Whenever a proposed BM step crosses the boundary, we just discard the proposed step and resample until the proposed move falls into the interior of $M$.

\subsection{Optimising the kernel hyper parameters and comparison with an RBF kernel in $\mathbb{R}$ } \label{optimT}

Given a diffusion time $t$,  using algorithm 1 we can generate a covariance matrix $\Sigma_{ {\text {\bf ff}} }^{t}$ for the training data indexed  by $t$. 
The log marginal likelihood function (over $f$) is given by \citep{Rasmussen2004}:
\begin{align}
\label{loglike}
p( \bm{y}| s) =  \int p( \bm{y}| {\text {\bf f} }) p({\text {\bf f} }|s) d{\text {\bf f} }   = -\frac{1}{2} \bm y^T (\Sigma_{ {\text {\bf ff}} }^{t} + \sigma_{noise}^2 I)^{-1} \bm y - \frac{1}{2} \log|\Sigma_{ {\text {\bf ff}} }^t +\sigma_{noise}^2I | - \frac{N_d}{2}\log2\pi.
\end{align}
%The term marginal likelihood refers to the marginalization over the function values {\bf f}.
The hyperparameters can be obtained by maximising the log of the marginal likelihood. The maximum of the BM diffusion time is set as $T*\Delta t$, where $T$ is a positive integer, and $\Delta t$ is the BM simulation time step as defined in \eqref{disProp}. 
% $\Delta t$ can be seen as the  `precision' of the smoothing parameter $t$. 
 $T$ covariance matrices $\Sigma_{ {\text {\bf ff}} }^{1...T}$ can be generated based on the BM simulations. Optimisation of diffusion time $t$ can be done by selecting the corresponding $\Sigma_{ {\text {\bf ff}} }^{t}$ that maximises the log marginal likelihood. Estimation of $\sigma_h$  given the smoothing parameter $t$ follows using standard optimisation routines, such as quasi-Newton.  For the sparse in-GP, the likelihood function is replaced by \eqref{sparseLik} and the hyper parameters can be obtained by similar procedures.
 % Following the similar procedure one  can estimate the kernel hyper parameters  by maximising the log of \eqref{sparseLik}.

We compare the estimates of  kernel hyperparameters from a normal GP and  the in-GP in $\mathbb R$ by applying both methods to ten sets of testing data.   Data sets are generated by sampling 20 data points from a multivariate normal distribution with mean zero and covariance $\Sigma_{test}$. $\Sigma_{test}$ is produced by a standard RBF kernel with $l=1$ and $\sigma_r =1$.  In this case, we know the ground truth of the hyperparameters of the heat kernel. 

We simulate $N_{bm}=40,000$ BM sample paths for each testing data point.  The estimates of hyper parameters $t$ and $\sigma_h$ are obtained by maximising \eqref{loglike}.  
  For the case of $\mathbb{R}$, the two methods should produce very similar results, since the true heat kernel is equivalent to a RBF kernel.
%\begin{align}
%\sigma_h^2 = \sigma_r^2 \cdot \sqrt{2\pi t}.
%\end{align}
The result is shown in Table \ref{sim.params} which records the true value and the median estimates of kernel hyper parameter $l$ and $\sigma$. Values in brackets show the median absolute deviation. The $p$-values of Wilcoxon tests indicate difference in medians between the two methods are not significant. 

\begin{table} \label{sim.params}
\caption{Comparison of estimates of kernel hyper parameters  from normal GP and in-GP in $\mathbb{R}$.}
\centering
\fbox{
%\centering
\begin{tabular}{*{6}{c}}
\em Case &\em Median estimates of $l$ &\em Median estimates of $\sigma_r$ \\
\hline
Trueth & 1& 1 \\
Normal GP& 1.13(0.16)&0.94(0.36) \\
in-GP   & 1.15(0.2) & 0.94(0.38)\\
p value &0.91 &0.85 \\
\end{tabular} 
} 
\end{table}

\section{Simulation studies} \label{simdata}

 In this section, we carry out simulation studies for a regression model with true regression functions defined on a U shape domain  and a 2-dimensional Swiss Roll embedded in $\mathbb R^3$. The performance of in-GP is compared  to that of a  normal GP   and  the soap film smoother in \cite{wood} for the U shape example. For the Swiss Roll example, the results from in-GP are compared with those from a normal GP model. 

\subsection{U shape example}

\begin{figure}[htp]
    \centering
    \subfigure[True function and data points]{\label{u:true}  \includegraphics[width=0.45\textwidth,height=0.45\textwidth]{True.pdf}}
     \subfigure[ \small{True function with less data points}]{\label{u16:true}  \includegraphics[width=0.45\textwidth,height=0.45\textwidth]{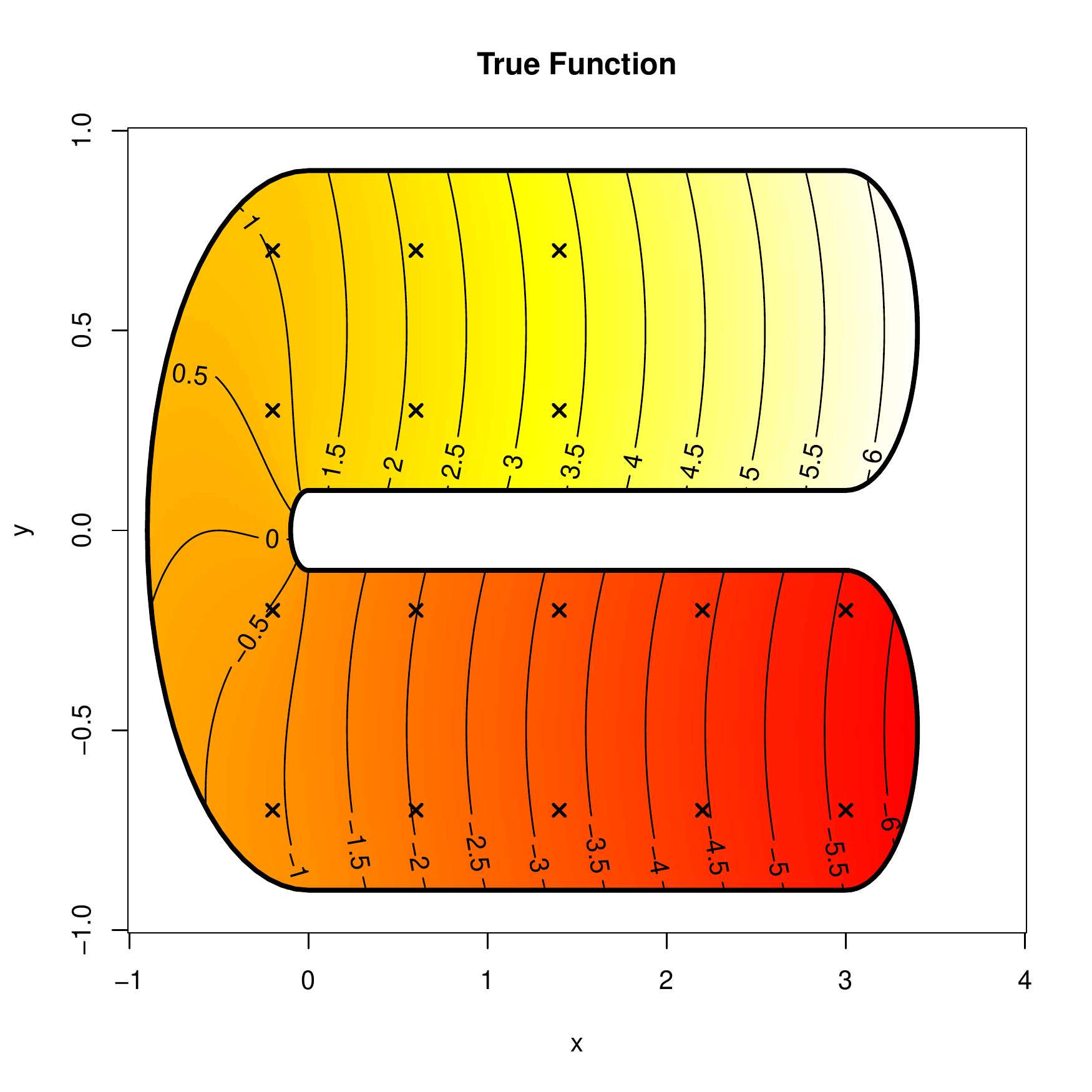}}
    \subfigure[GP prediction with all data points]{\label{u:gp}  \includegraphics[width=0.45\textwidth,height=0.45\textwidth]{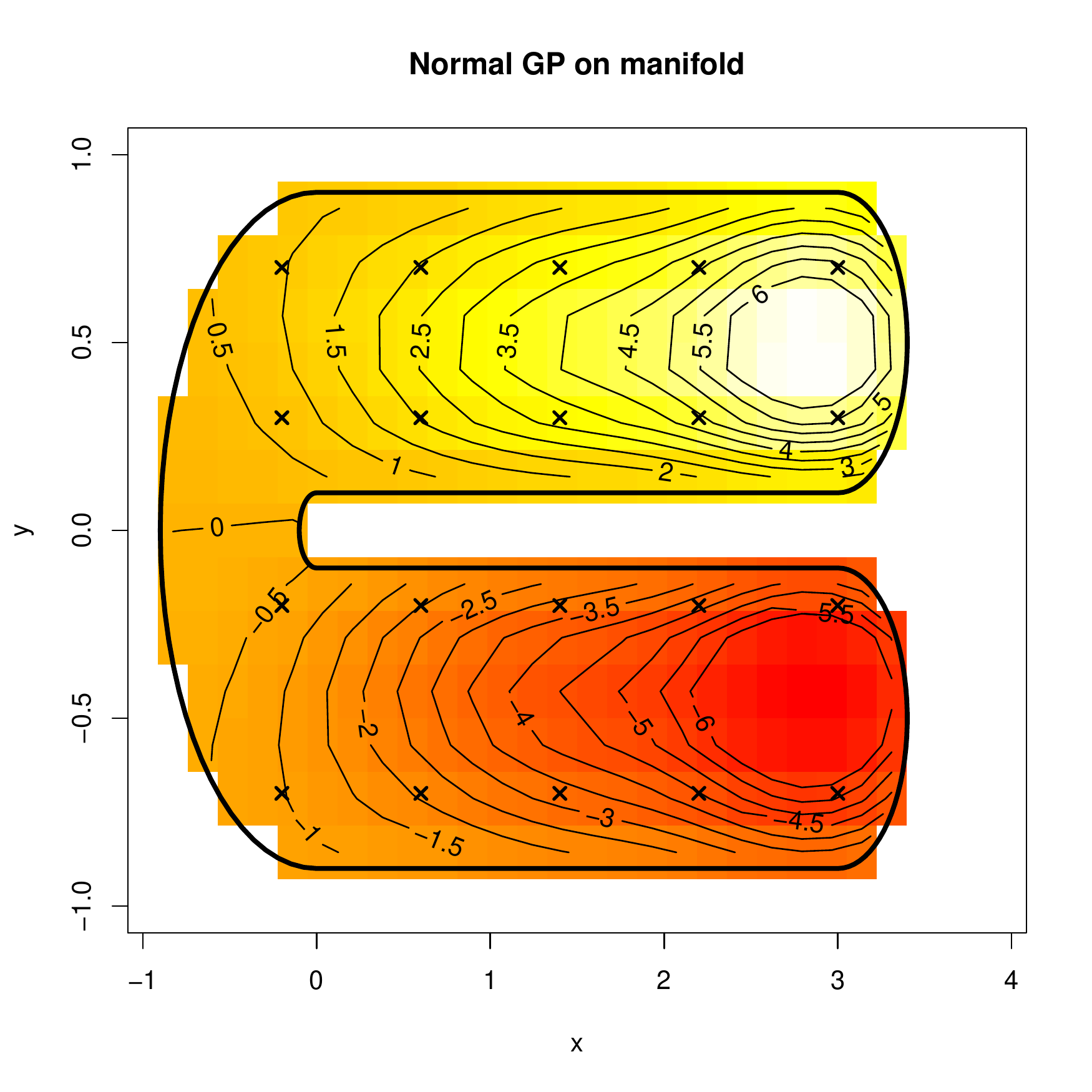}}
    \subfigure[GP prediction with less data points]{ \label{u16:gp} \includegraphics[width=0.45\textwidth,height=0.45\textwidth]{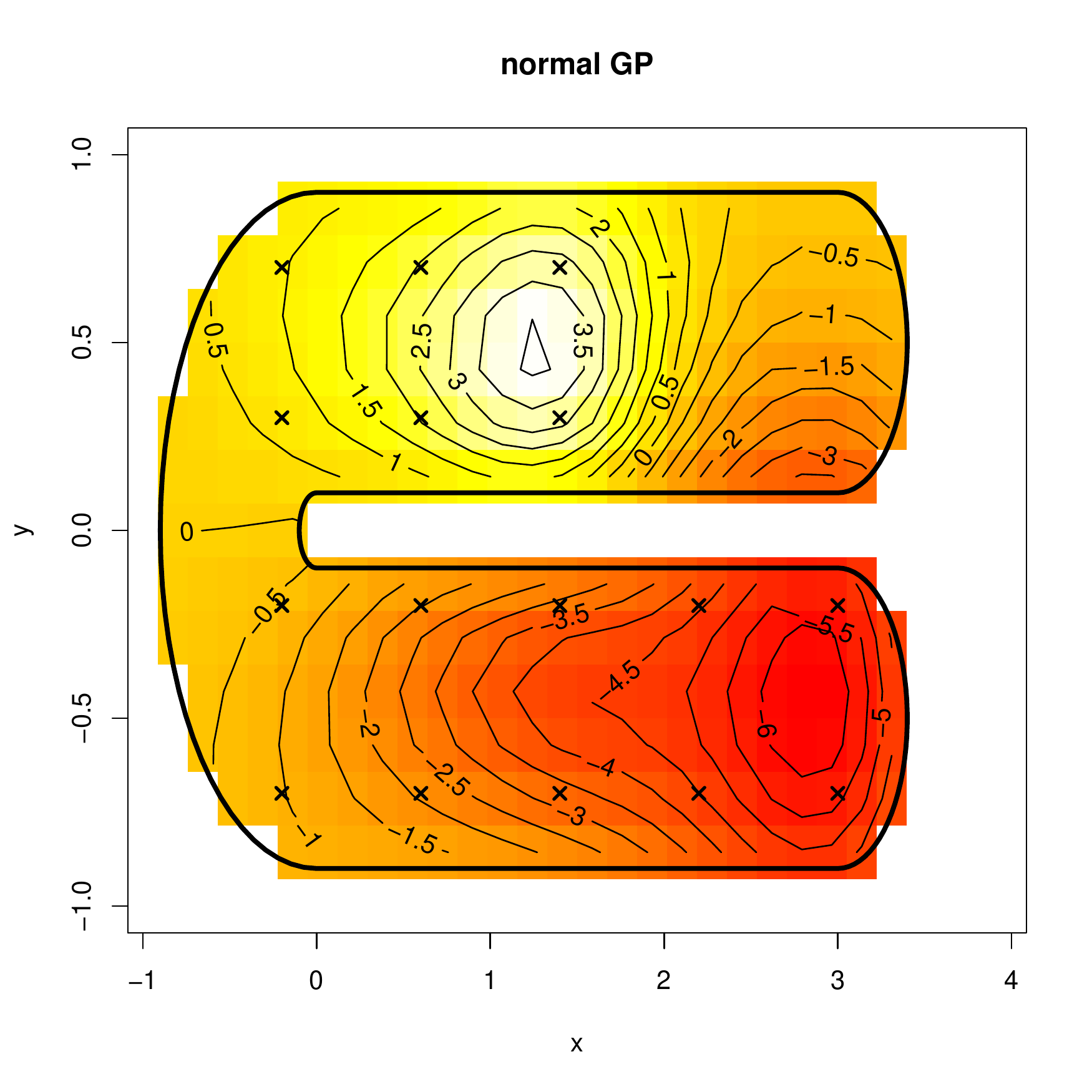}}
    \subfigure[in-GP prediction with all data points]{\label{u:gpm}  \includegraphics[width=0.45\textwidth,height=0.45\textwidth]{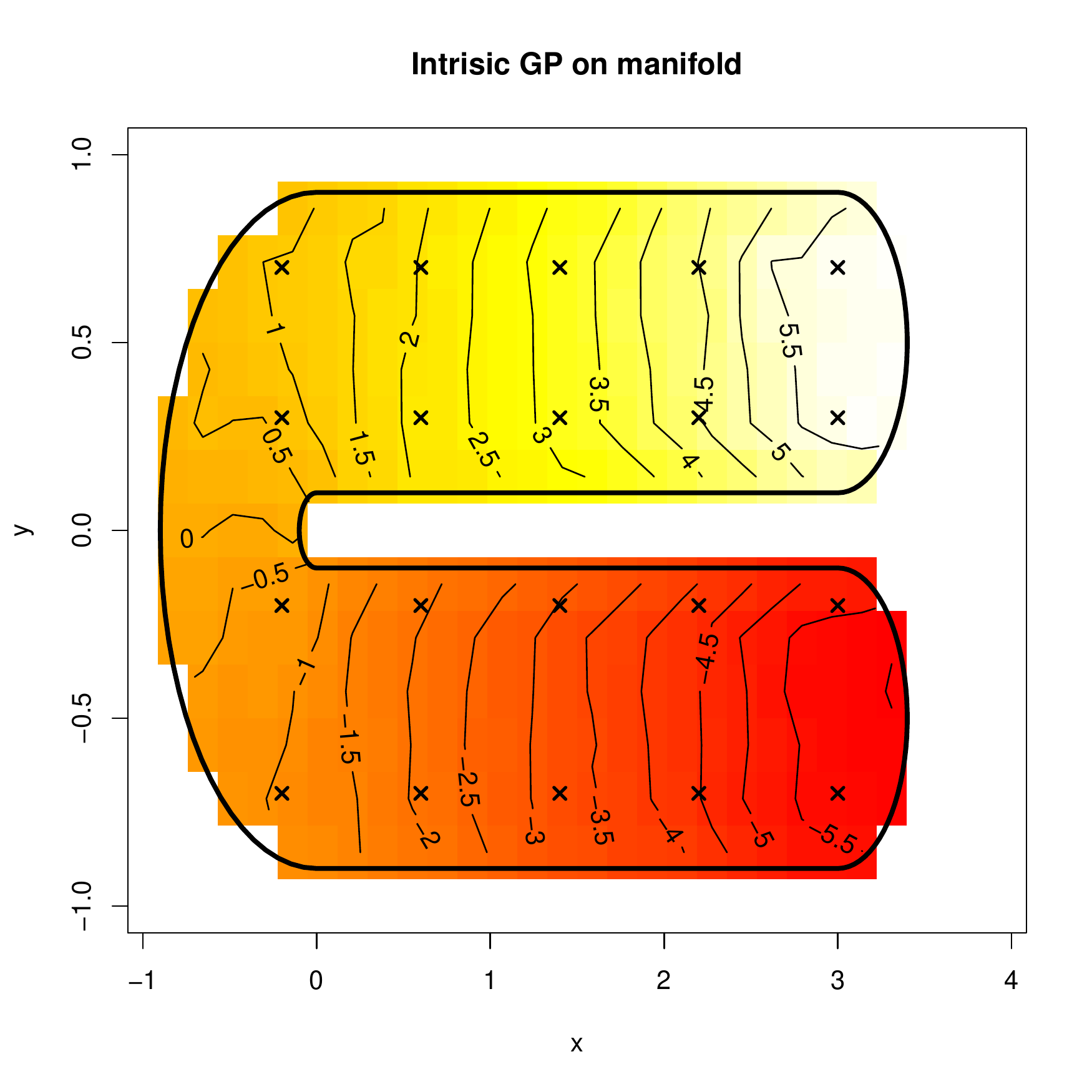}}
    \subfigure[in-GP prediction with less data points]{\label{u16:gpm} \includegraphics[width=0.45\textwidth,height=0.45\textwidth]{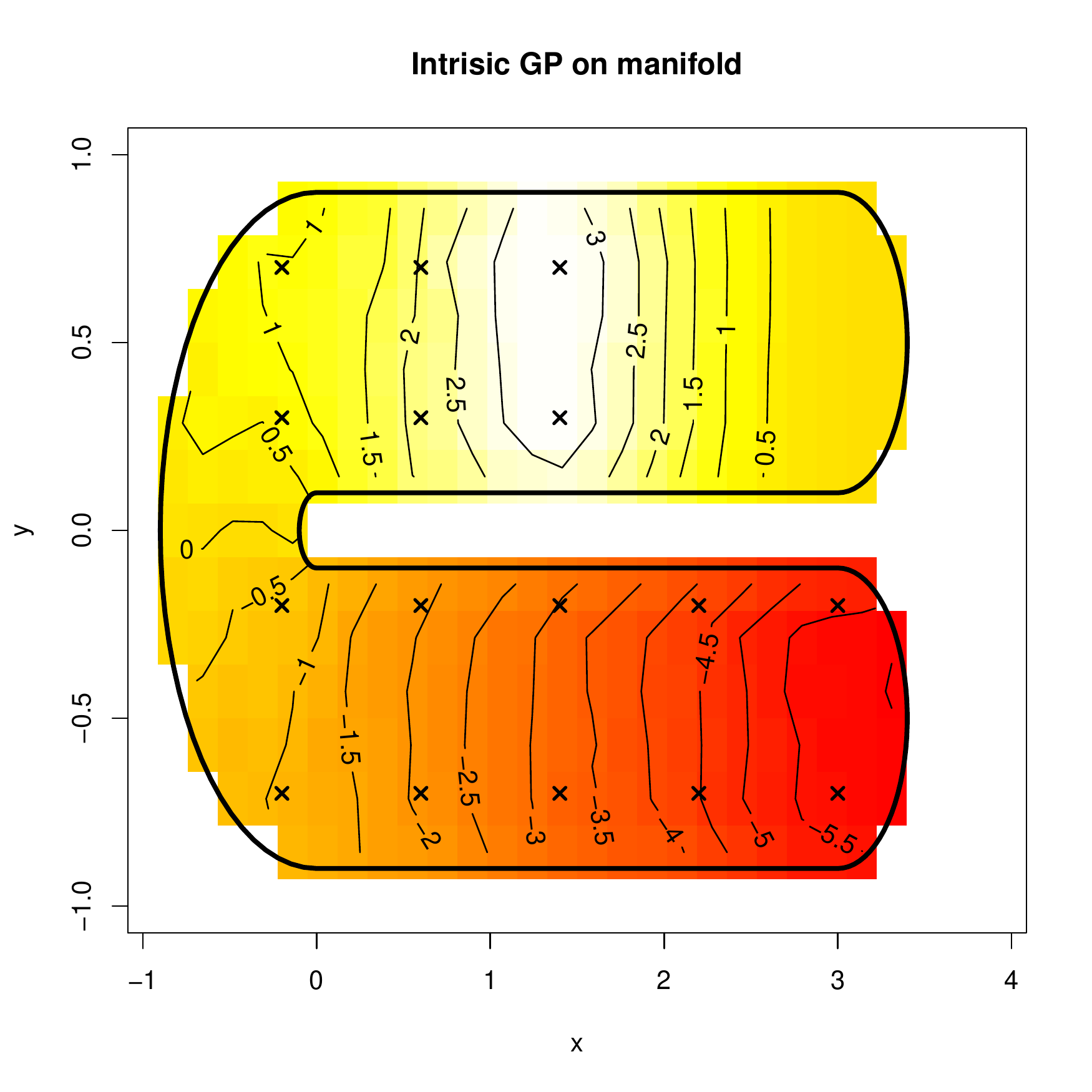}}
    \caption{\label{fig:Ushape}
    \footnotesize
    { Comparison of in-GP  and normal GP in the U-shape example.}  }
\end{figure}

A U shaped domain  (see e.g., \cite{wood2001}), defined as a subset of $\mathbb R^2$ is plotted in Figure \ref{u:true}. The value of a test or regression function  (i.e. the color of the map) varies smoothly from the  lower right corner to the upper right corner of the  domain ranging from -6 to 6. The black crosses represent 20 observations which were equally spaced  in both $x$ and $y$ directions within the domain of interest. The goal is to estimate the test function and make predictions at 450 equally  spaced grid points within the domain. 

\begin{figure}[htb]
 \centering
 \includegraphics[width=0.8\textwidth,height=0.4\textwidth]{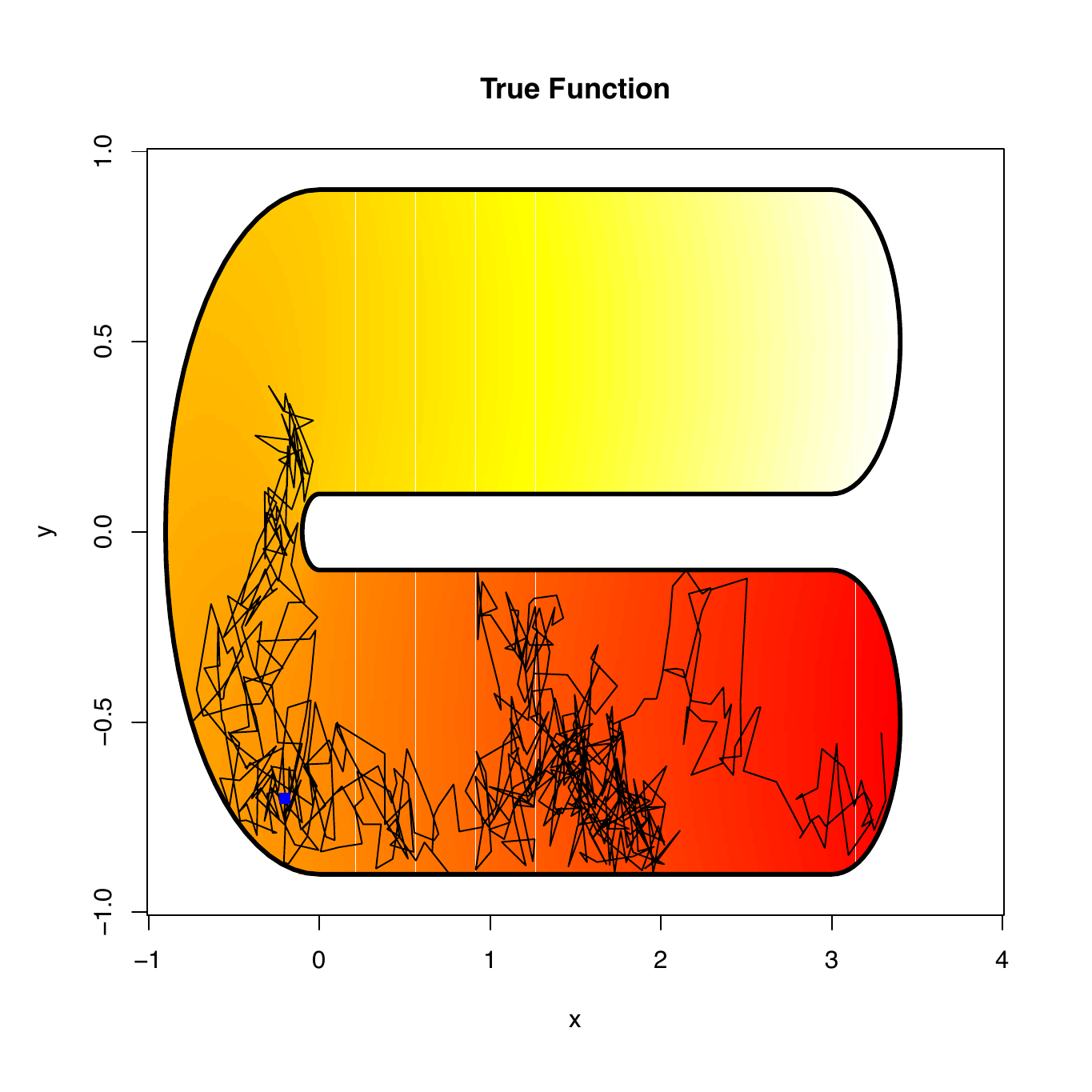}
    \caption{\label{fig:Ushape}
    \footnotesize
    { \label{u:path} A sample path of BM on the U-shape domain  }}
\end{figure}

Since the U shaped domain is defined as a subset of $\mathbb R^2$, the mapping function $\phi$ in equation \eqref{fig:metric} is a constant. Therefore, BM reduces to the  standard BM in the two dimensional Euclidean space restricted within the boundary. When a proposed BM step hits the boundary, the Neumann boundary condition is applied and  the proposed move is rejected. New proposal steps will be made until the proposed  sample path locates within the boundary. The trajectory of a sample path (black line) of the BM  is shown in Figure \ref{u:path} with the blue dot serving as the starting location of the BM.

The heat map of the predictive mean of in-GP at the grid points is shown in Figure \ref{u:gpm}. The colored contours of the prediction are similar to that of the true function in Figure \ref{u:true}.  The contours of the normal GP predictive mean in Figure \ref{u:gp} are more squashed,  and the differences are  exacerbated when certain observations  are removed as in Figure \ref{u16:true}. It is clear that the normal GP smooths across the gap between the two arms of the domain (see Figure \ref{u16:gp}). This is due to that fact that the upper arm and lower arm are close in Euclidean distance.  In contrast, the in-GP, which takes into account the intrinsic geometry, does not smooth across the gap as seen in Figure \ref{u16:gpm}.  Given a fixed diffusion time, the transition probability of BM from points in the lower arm to points in the upper arm within the boundary is relatively small. This leads to lower covariance between these two regions and more accurate predictions.

The U shaped domain  example has also been used for evaluating the performance of the soap film smoothers in \cite{wood}, in comparison with some other methods such as  thin plate splines and the FELSPLINE method \citep{tramsay}.  Comparisons made in \cite{wood} show that the soap film smoother outperforms the other two methods. In our study,  the  in-GP, normal GP, and soap film smoother are compared by varying different levels of signal-to-noise ratio. The values of the true function are perturbed by Gaussian noise with a standard deviation of $0.1$ and $1$ (signal-to-noise ratios  are 30db and 10db, respectively) with 50 replicates for each noise level. For each of the replicates, different methods are  applied to estimate the test function at grid points. The mean and standard deviation of the mean-squared error (MSE) for these 50 replicates are reported in table \ref{sim.u}. 
The soap film smoother is constructed using 10 inner knots and 10 cubic splines. in-GP and soap film are both significantly better than normal GP, but there is no substantial difference between the two methods.

\begin{table} \label{sim.u} 
\caption{Comparison of the rms error of predictive means for different methods. The table shows the mean of rms error over 50 datasets.  Values in bracket show the standard deviation.}
\centering{
\fbox{
\begin{tabular}{*{6}{c}}
\em Case &\em normal GP & in-GP & soap film smoother \\
\hline
30db & 1(0.01)& 0.274(0.04) &0.271(0.22) \\
10db & 1.36(0.17)& 0.754(0.14) &0.747(0.37) \\[1ex]
\end{tabular}} }
\end{table}

\subsection{Swiss Roll}

The in-GP model  applies to general Riemannian manifolds and has  much wider applicability to complex spaces beyond  subsets of $\mathbb R^2$. Here we consider a synthetic dataset on a Swiss Roll which is two-dimensional manifold embedded in $\mathbb R^3$. The soap film method is only appropriate for smoothing over regions of $\mathbb R^2$, and hence cannot be applied here. 

A Swiss Roll is a spiralling band in a three-dimensional  Euclidean space. A nonlinear function $f$ is defined on the surface of Swiss Roll with
\begin{align*}
Y_i= f( x_i,y_i,z_i ) + \epsilon_i,
\end{align*}
where $x_i,y_i,z_i $ are the coordinates of a point on the surface. The construction of the Swiss Roll and the derivation of the metric tensor are shown in the Appendix.  

\begin{figure}[htp]
    \centering{
    \subfigure[True function on Swiss Roll]{\includegraphics[width=0.45\textwidth,height=0.45\textwidth]{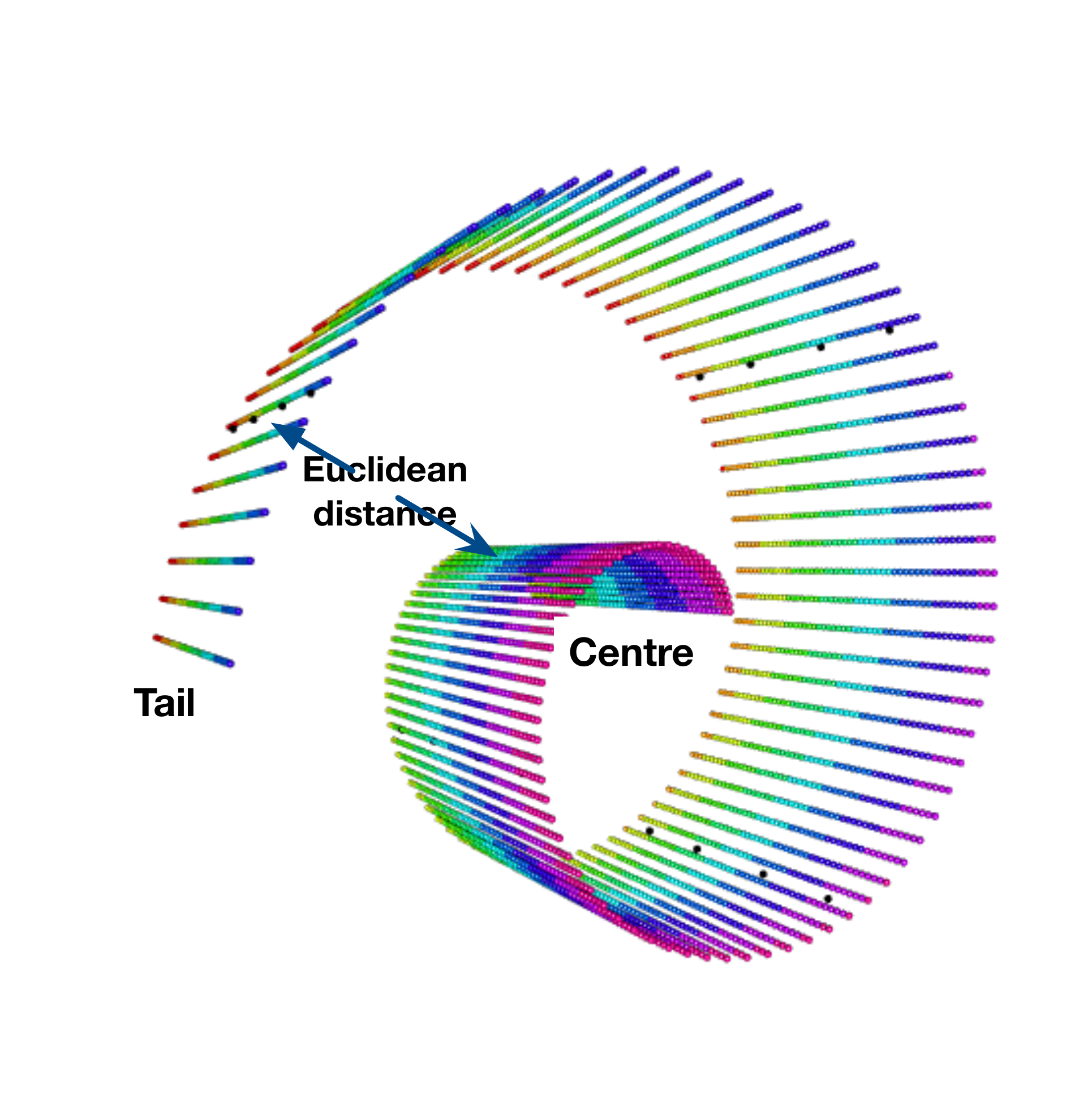}}
    \subfigure[Unfolded true function and data]{\includegraphics[width=0.45\textwidth,height=0.45\textwidth]{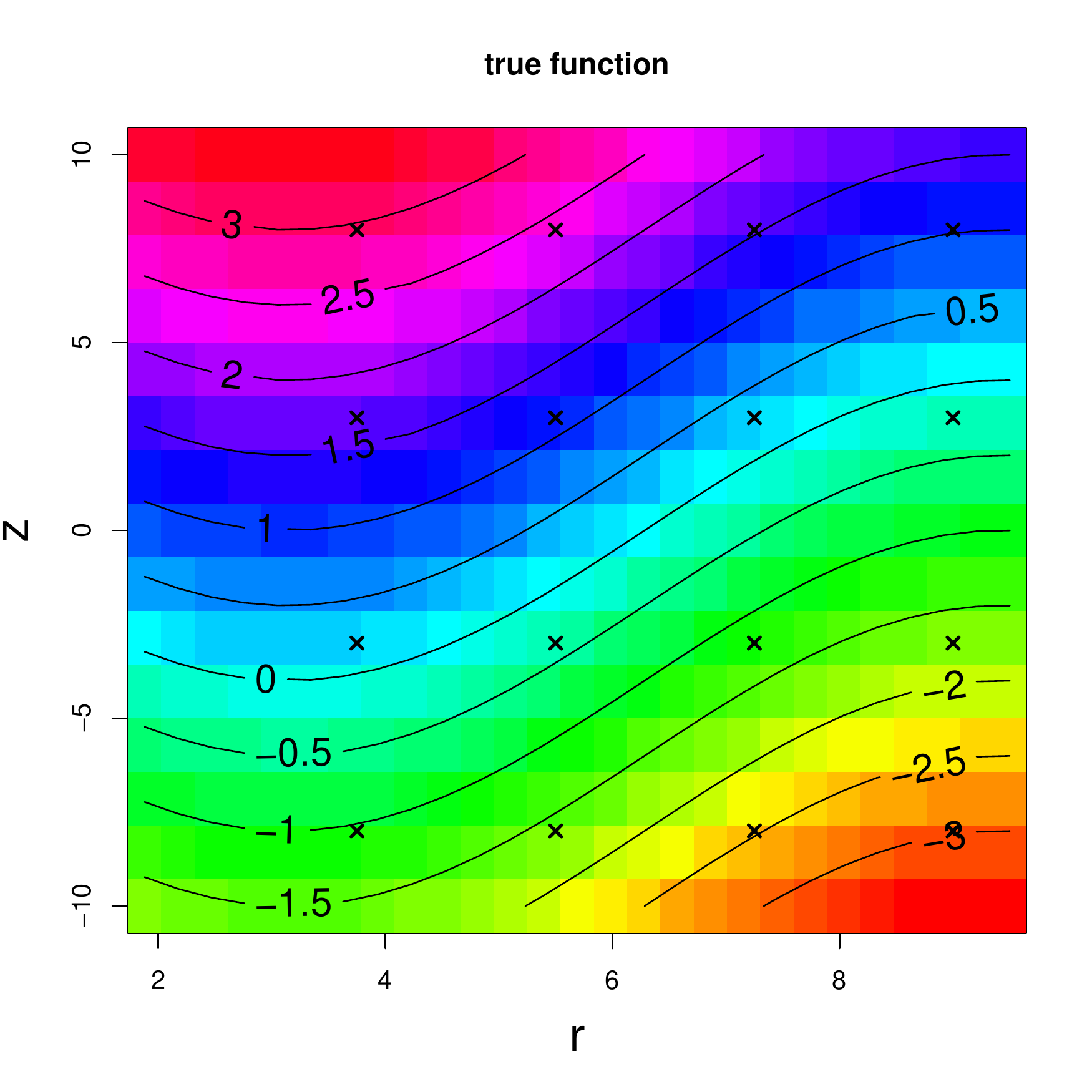}}
   % \subfigure[2D views of Swiss Roll]{\includegraphics[width=0.45\textwidth,height=0.45\textwidth]{2dswiss.pdf}}
    \subfigure[normal GP prediction]{\includegraphics[width=0.45\textwidth,height=0.45\textwidth]{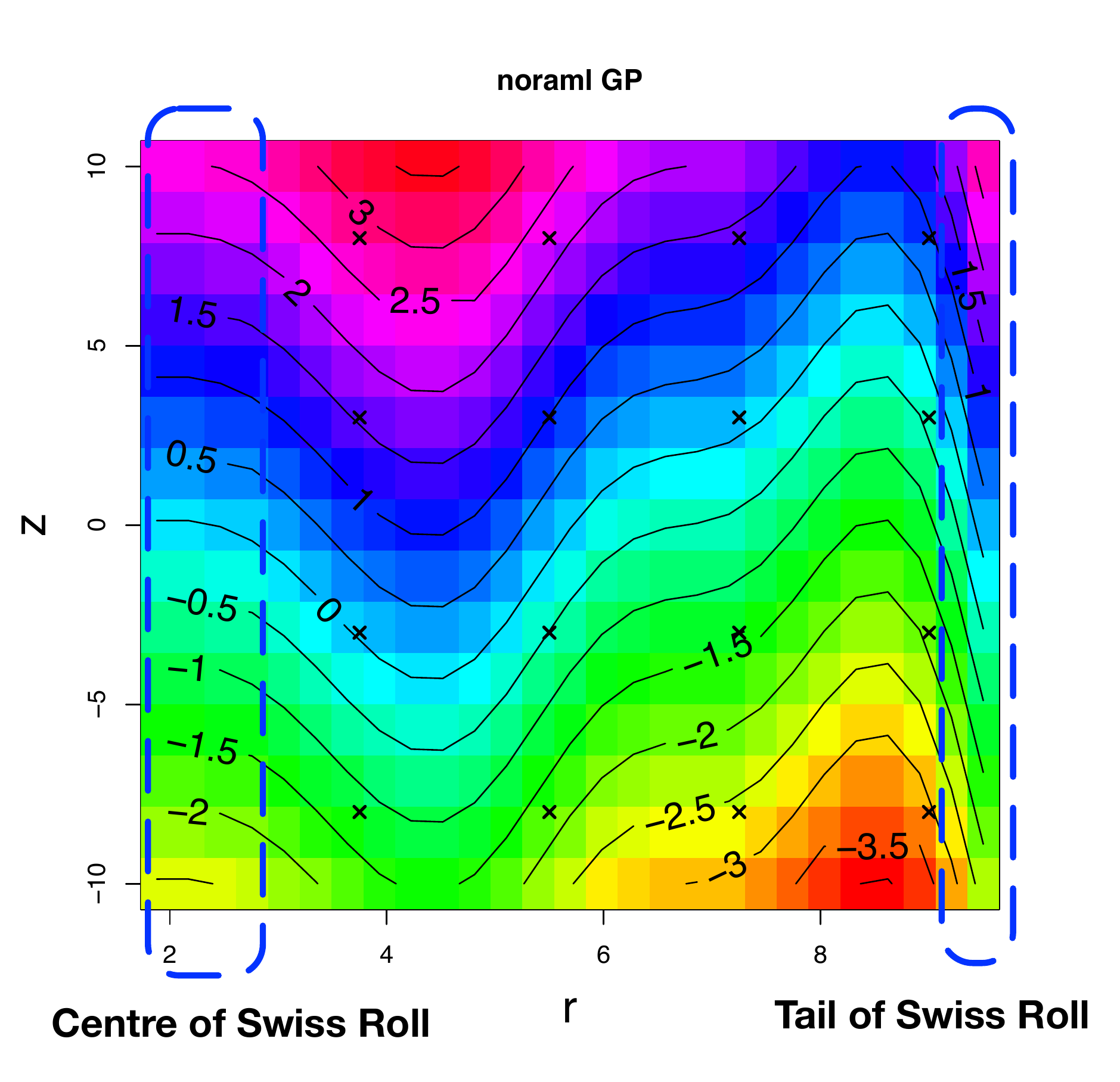}}
   % \subfigure[BM on Swiss Roll]{\includegraphics[width=0.45\textwidth,height=0.45\textwidth]{BMswiss.pdf}}
    \subfigure[in-GP prediction]{\includegraphics[width=0.45\textwidth,height=0.45\textwidth]{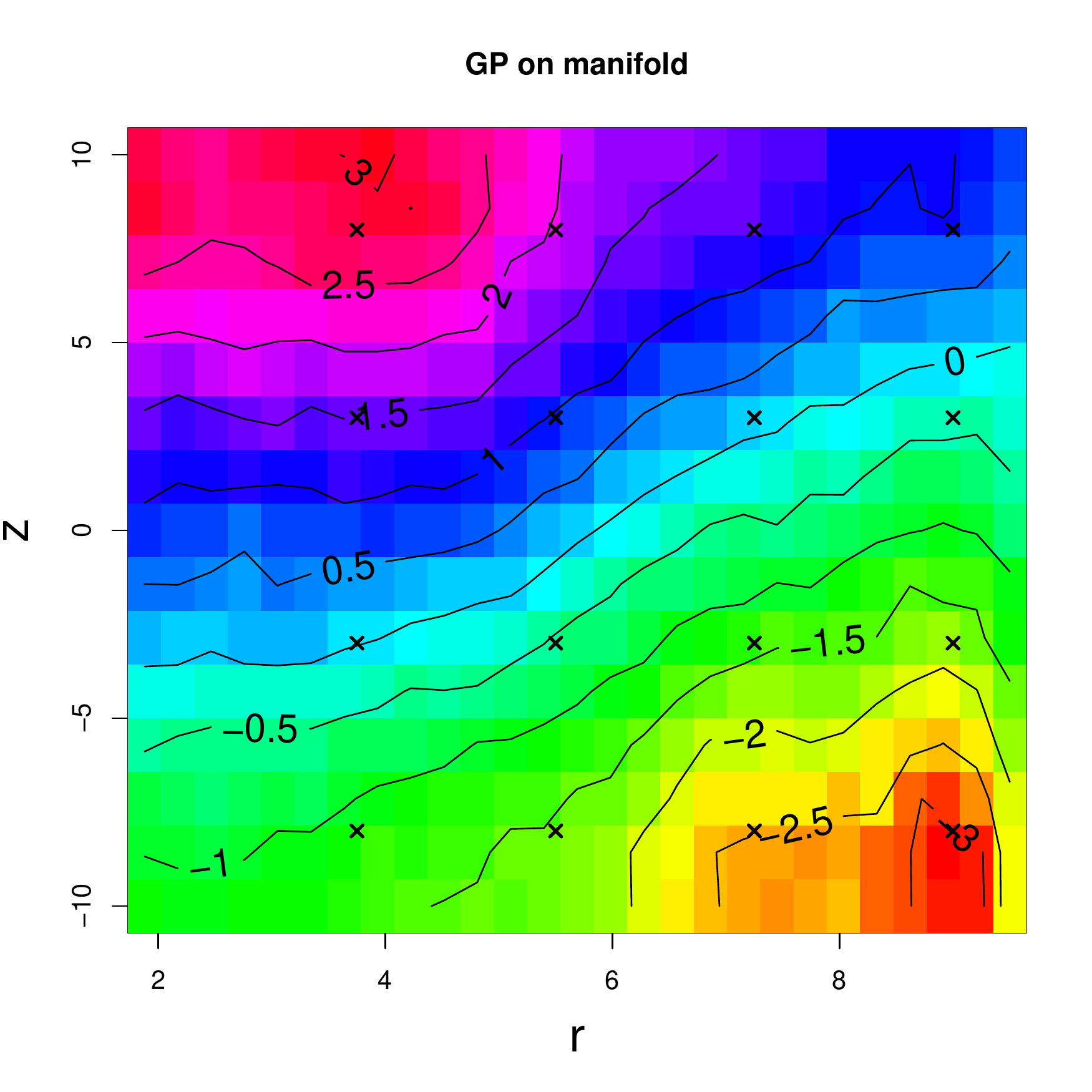}}
    \caption{\label{fig:swiss}
    \footnotesize
    { Comparison of in-GP and  normal GP on Swiss Roll.}  } }
\end{figure}

 The true function $f$ is plotted in Figure \ref{fig:swiss}(a). 20 equally spaced observations are marked with black crosses.  For better visualisation, the true function is plotted in the unfolded Swiss Roll in  the radius $r$ and width $z$ coordinates in Figure \ref{fig:swiss}(b). The true function values are indicated by colour and with contours at the grid points. 

We first applied a normal GP to this example using an RBF kernel in $\mathbb R^3$. In order to visualise the differences between the prediction and the true function, the GP predictive mean is plotted in the unfolded Swiss Roll in Figure \ref{fig:swiss}(c).  The overall shape of contours is more wiggly comparing to the true function in Figure \ref{fig:swiss}(b).   In addition, the predictive mean is quite different from the truth in colour in certain regions.  For example, the left end of Figure \ref{fig:swiss}(c)  marked by the blue dashed box corresponding to the centre of the Swiss Roll and the right end of Figure \ref{fig:swiss}(c) corresponding to the tail of the Swiss Roll.  The prediction performance of the normal GP in these regions is poor. This is because the Euclidean distance between the two regions is small as shown in Figure \ref{fig:swiss}(a) while the the geodesic distance between them (defined on the surface of Swiss Roll) is big. 
%The standard RBF kernel with Euclidean distance in $\mathbb R^3$ assigns high covariance between the two regions resulting in  poor predictions.

\begin{figure}[htp]
 \centering
 \includegraphics[width=0.5\textwidth,height=0.5\textwidth]{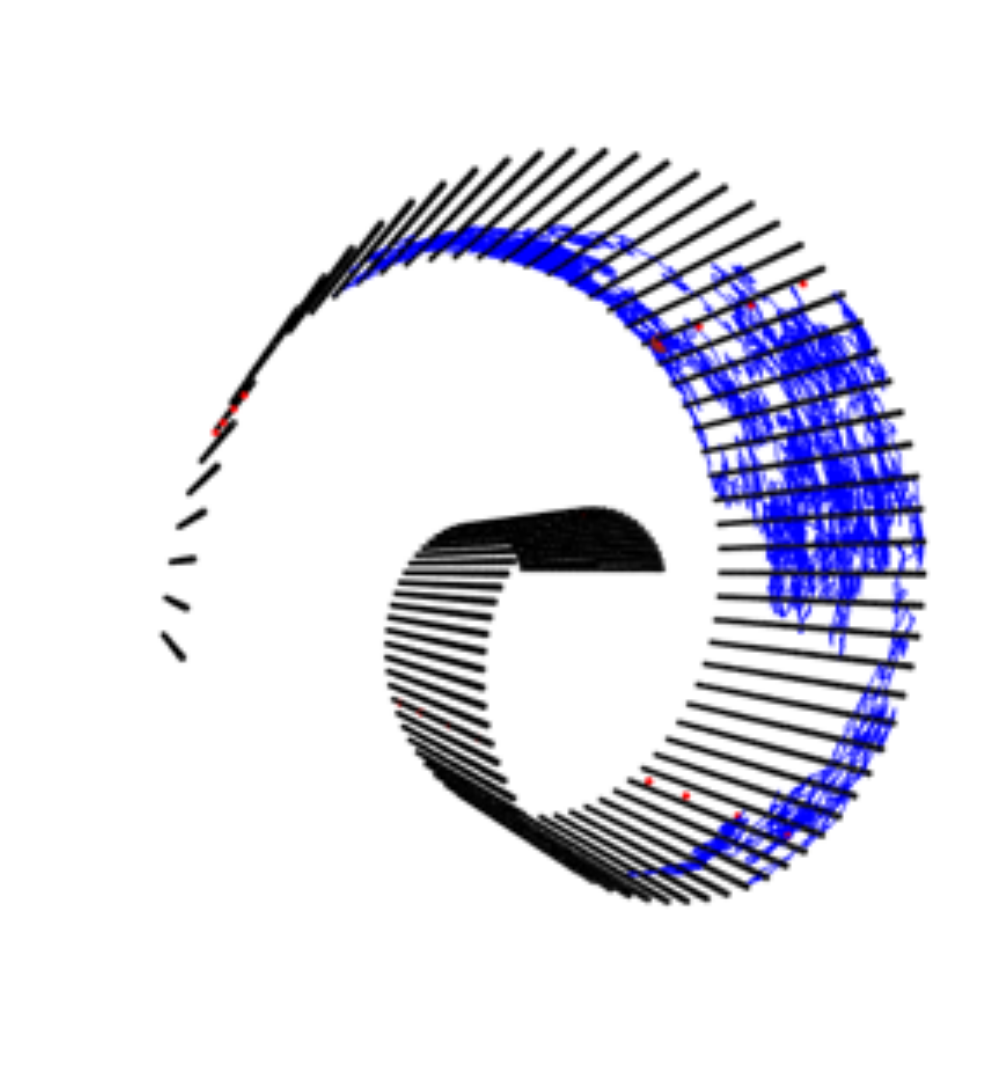}
    \caption{\label{fig:Ushape}
    \footnotesize
    { \label{fig:swissBM} A sample path of BM on the Swiss Roll.}}
\end{figure}

In applying the in-GP to these data,  the BM sample paths can be simulated using equation \eqref{disBM} using  the metric tensor of the Swill Roll.  In particular, BM on the Swiss Roll can be modelled as the stochastic differential equation:
\begin{align}
dr(t) &= \frac{-2r}{(1+r^2)^2} dt + \frac{1}{2} \frac{2r}{(1+r^2)^2} dt + (1+r^2)^{-1/2} dB_r(t),\\
dz(t) &=  dB_z(t),
\end{align}
where $B_r(t)$ and $B_z(t)$ are two independent BMs in Euclidean space. A trace plot of a single BM sample path is shown in Figure \ref{fig:swissBM}. Following the procedure introduced in section \ref{GPmani}, the predictive mean of in-GP is shown in Figure \ref{fig:swiss}(d). The overall shape of the contour of the predictive mean is similar to that of the true function. The prediction at the centre and tail part of the Swiss Roll has been improved comparing to the results of the normal GP. 
%The improvement can be explained by the nature of our method. The BM paths travel longer distance from the centre to the tail part of Swiss Roll. Thus the transition density is relatively small which leads to smaller covariance between the two regions. 
The root mean square error is calculated between the predictive mean and the true value at the grid points. It has been reduced from $0.53$ (normal GP) to $0.29$ for the in-GP. 
%\vspace{-3mm}

\section{Application to chlorophyll data in Aral sea} \label{real}
\begin{figure}
    \centering
    \subfigure[chlorophyll data in Aral sea]{\label{aral:data} \includegraphics[width=0.45\textwidth,height=0.45\textwidth]{araldata.pdf}}
        \subfigure[chlorophyll data with west basin removed]{\label{aral:datal} \includegraphics[width=0.45\textwidth,height=0.45\textwidth]{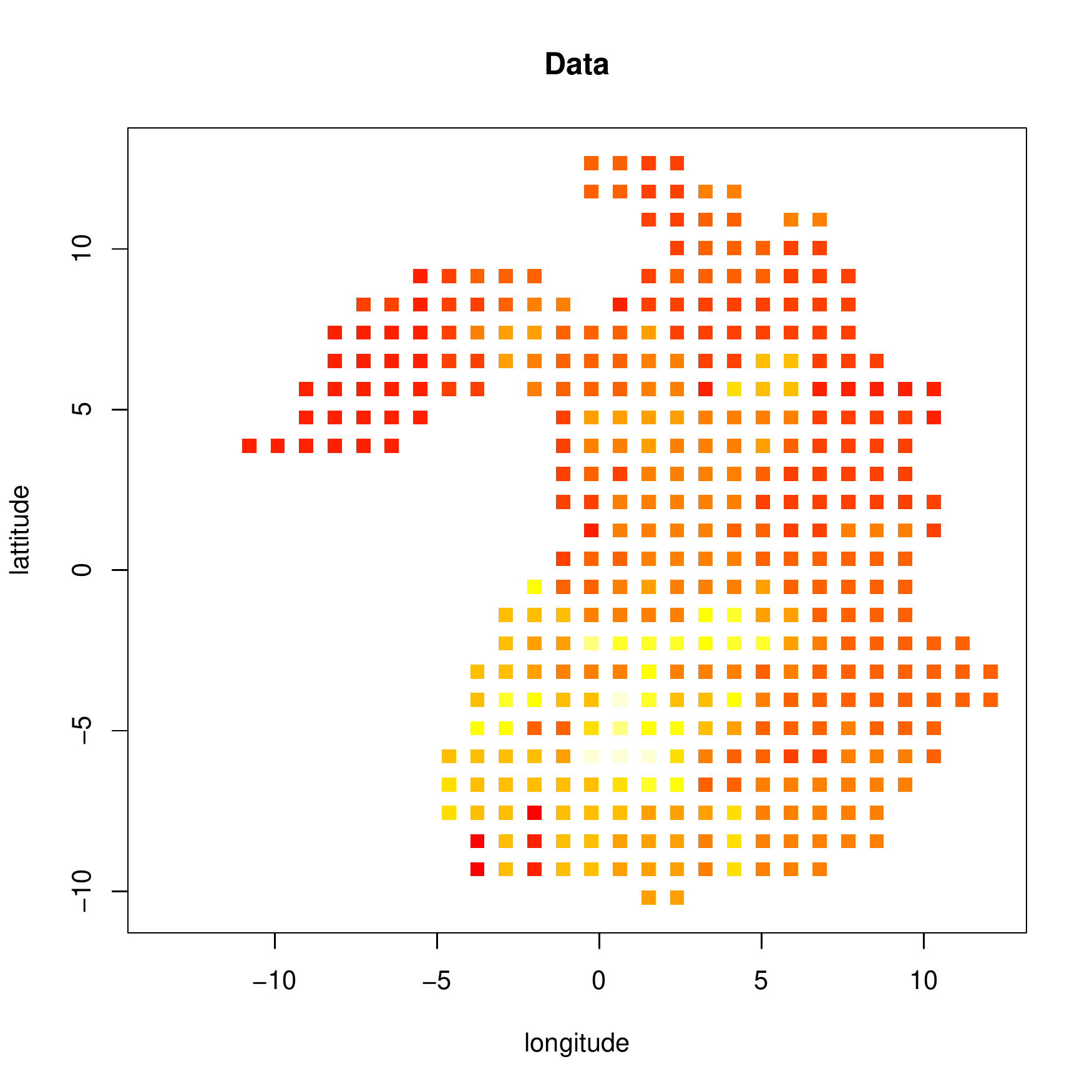}}
            \subfigure[GP prediction]{\label{aral:gp} \includegraphics[width=0.45\textwidth,height=0.45\textwidth]{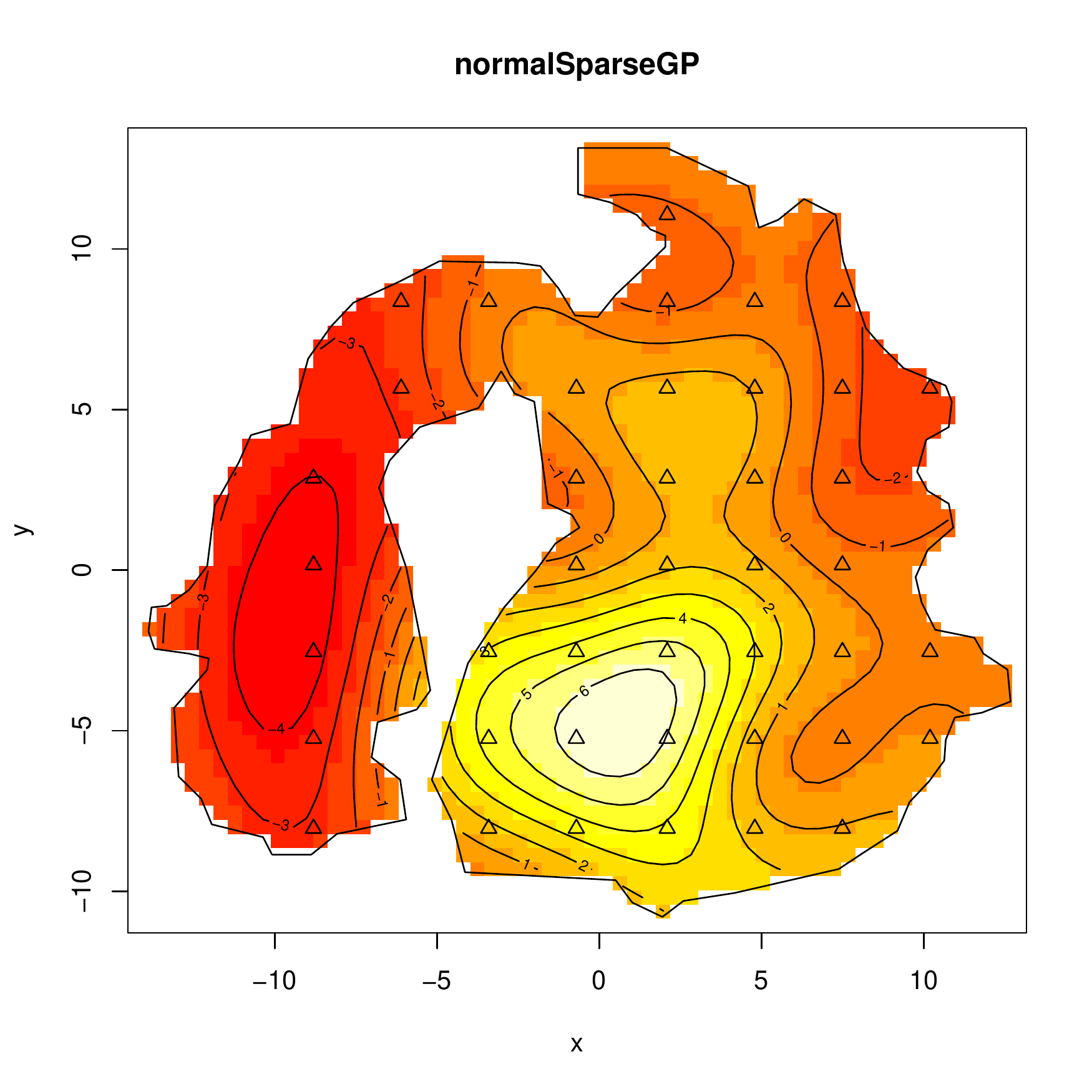}}
    \subfigure[GP prediction with less data]{\label{aral:gpl} \includegraphics[width=0.45\textwidth,height=0.45\textwidth]{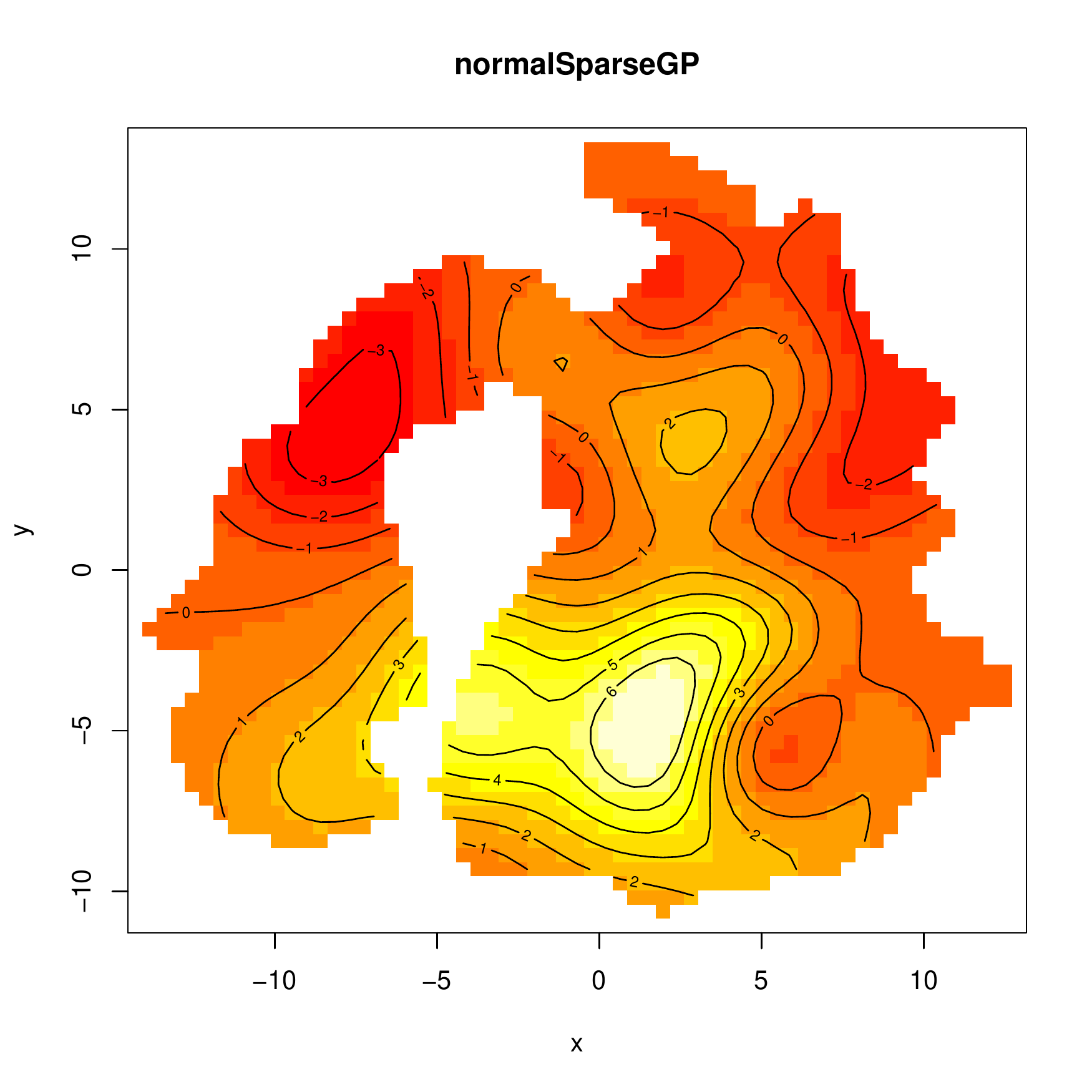}}
            \subfigure[sparse in-GP prediction]{\label{aral:gpmani} \includegraphics[width=0.45\textwidth,height=0.45\textwidth]{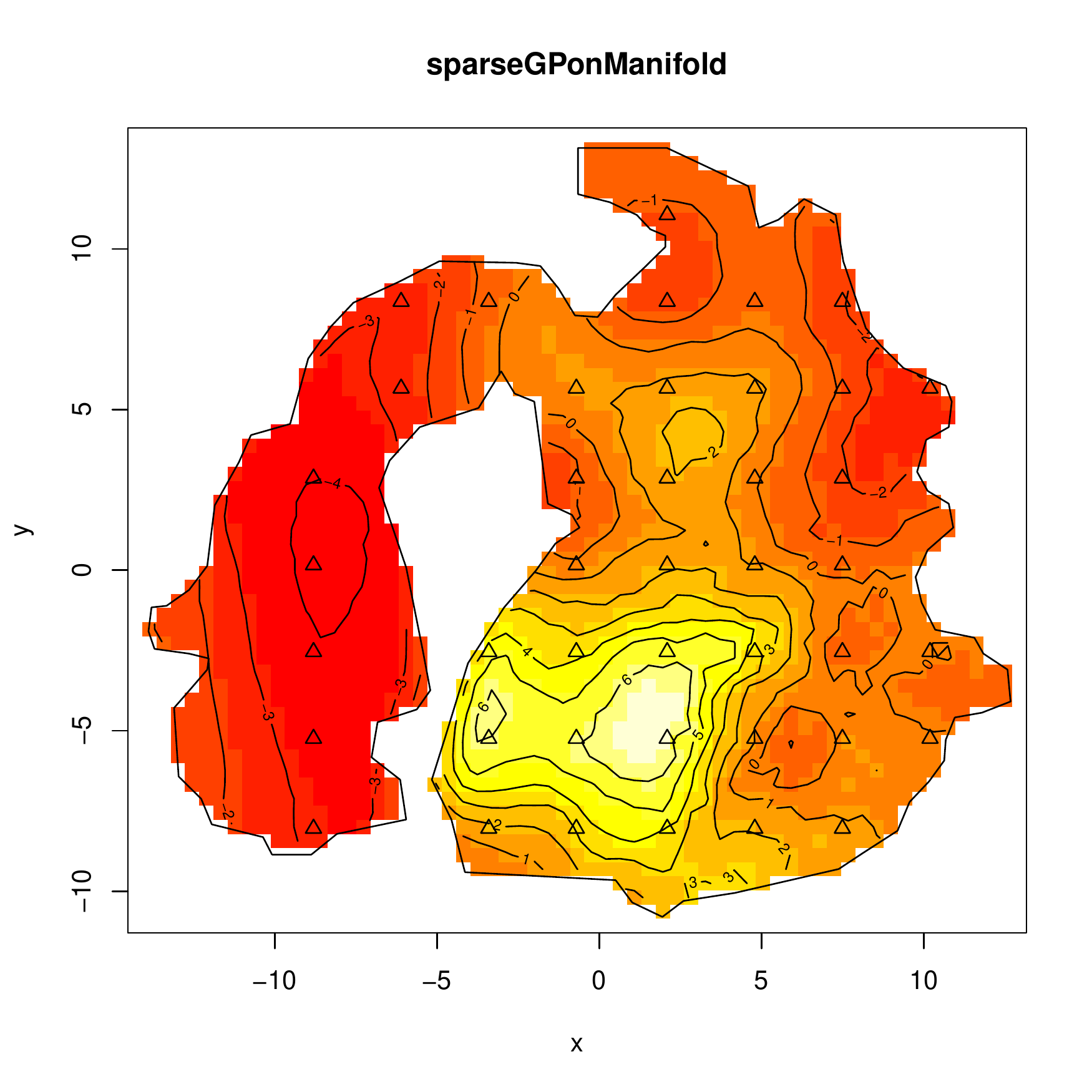}}
    \subfigure[sparse in-GP prediction with less data]{\label{aral:gpmanil} \includegraphics[width=0.45\textwidth,height=0.45\textwidth]{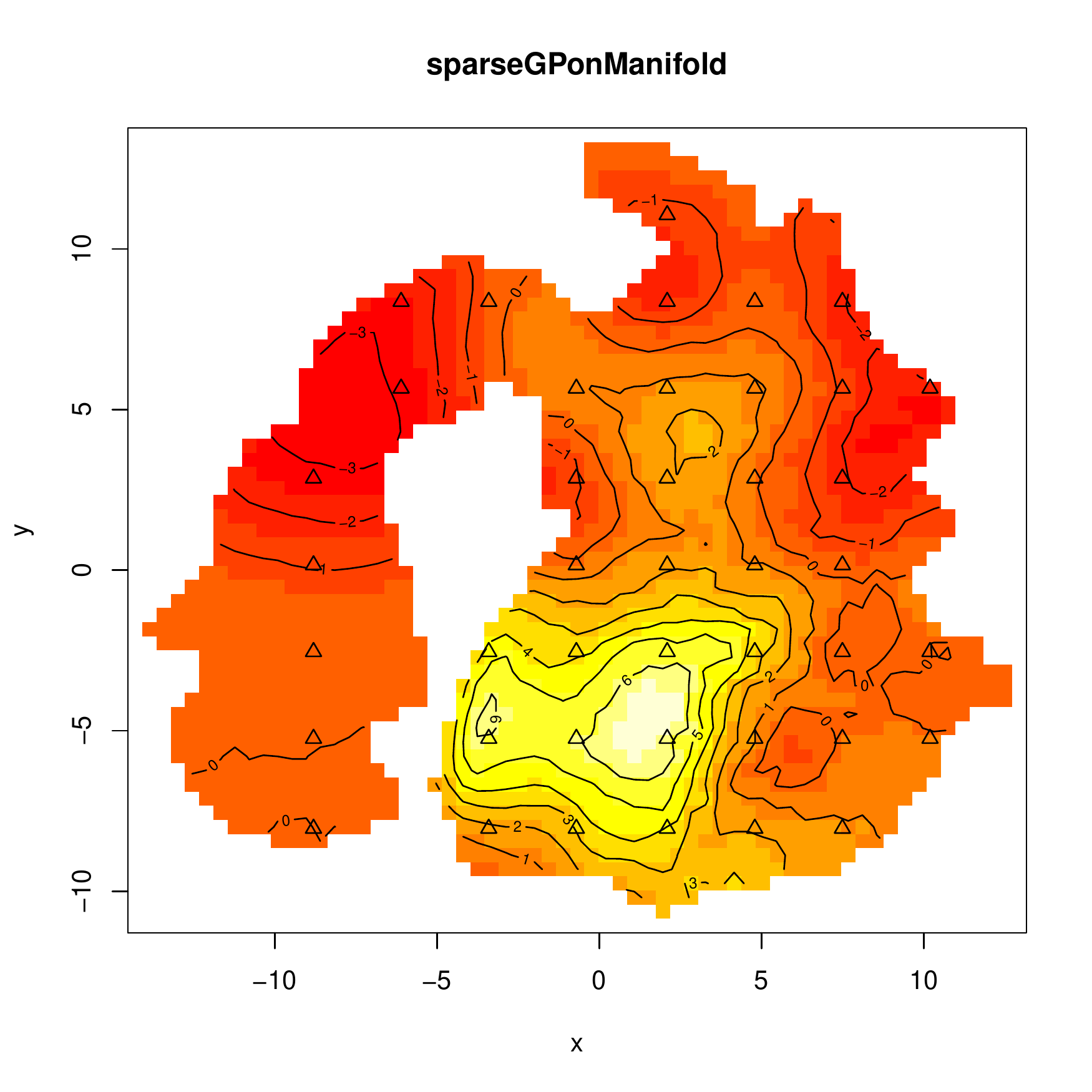}}

    \caption{
    \footnotesize
    { Comparison of in-GP and  a normal GP for chlorophyll data in Aral sea.}  }
\end{figure}

In this section, we consider an analysis of remotely-sensed chlorophyll data at $485$ locations in the Aral sea.  The data are available from  the gamair package \citep{wood2016} and are plotted in Figure \ref{aral:data}. The level of chlorophyll concentration is represented by the intensity of the colour. The chlorophyll  data from the satellite sensors are noisy and vary smoothly within the boundary but not across the gap corresponding to the isthmus of the peninsula. We applied different methods to estimate the spatial pattern of the chlorophyll density.

The $log$ of chlorophyll concentration is modelled as a function of the latitude and longitude coordinates of the measurement locations:
\begin{align*}
chl_i = f(lon_i, lat_i) + \epsilon_i,
\end{align*}
where $lon_i$ and $lat_i$ are standardised by removing the mean of the longitude and latitude from the raw coordinates.

In order to reduce the computation cost from simulating BM sample paths starting at all the observed points and grid points, the sparse in-GP model from section \ref{sec:sparseGP} is applied. 42 inducing points are introduced that are equally spaced within the boundary of the Aral sea. The inducing points are represented by small triangles in Figure \ref{aral:gpmani}. The required number of BM sample paths has been reduced from $485 \times N_{BM}$ to $42 \times N_{BM}$, where $N_{BM}$ is 20,000 in this example. 

\begin{figure}
\centering
 \subfigure[Predictive variance] {\label{fig:var}
  \includegraphics[width=0.45\textwidth,height=0.45\textwidth]{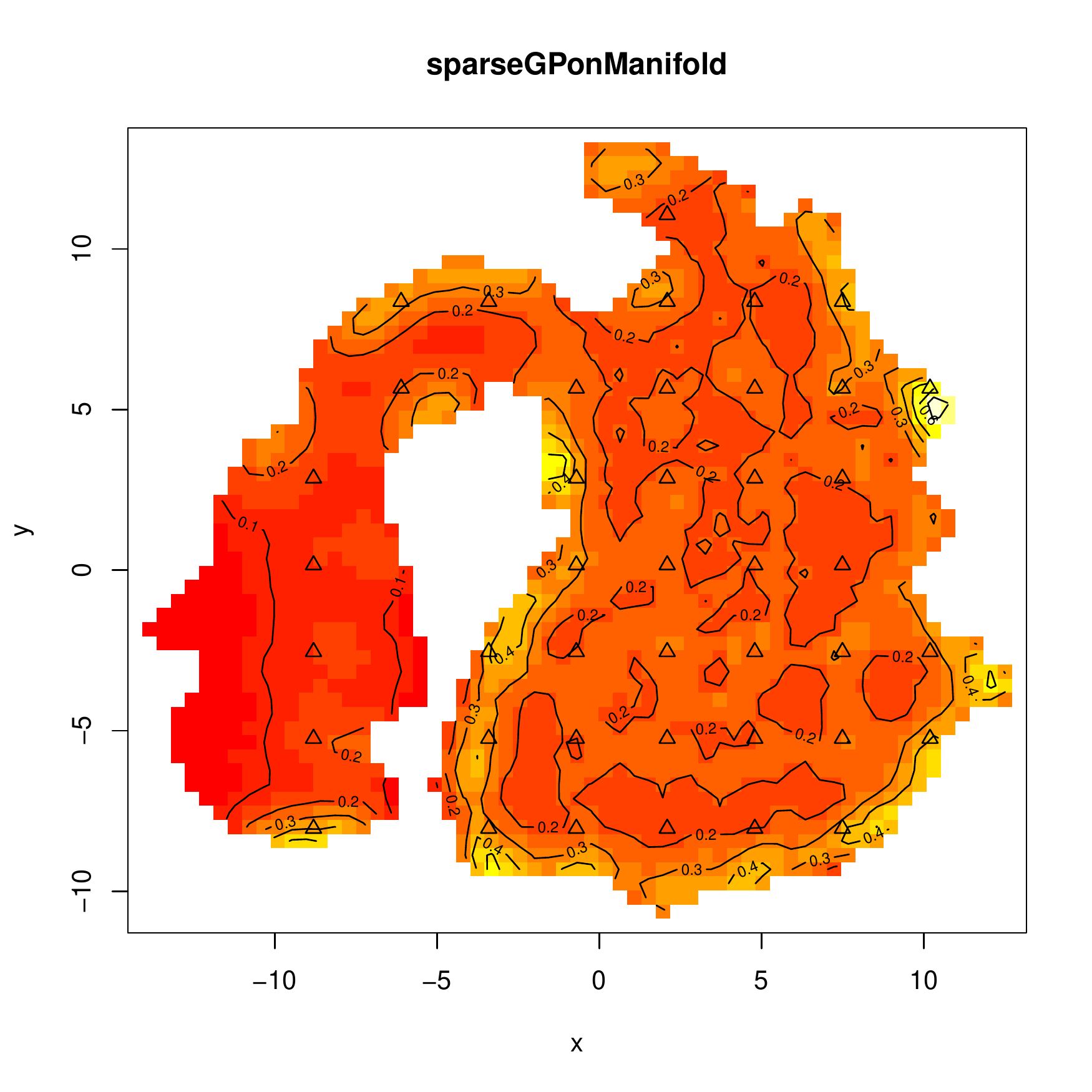} }
  \subfigure[Predictive variance with data points removed from west basin of the sea ] {\label{fig:varleft}
\includegraphics[width=0.45\textwidth,height=0.45\textwidth]{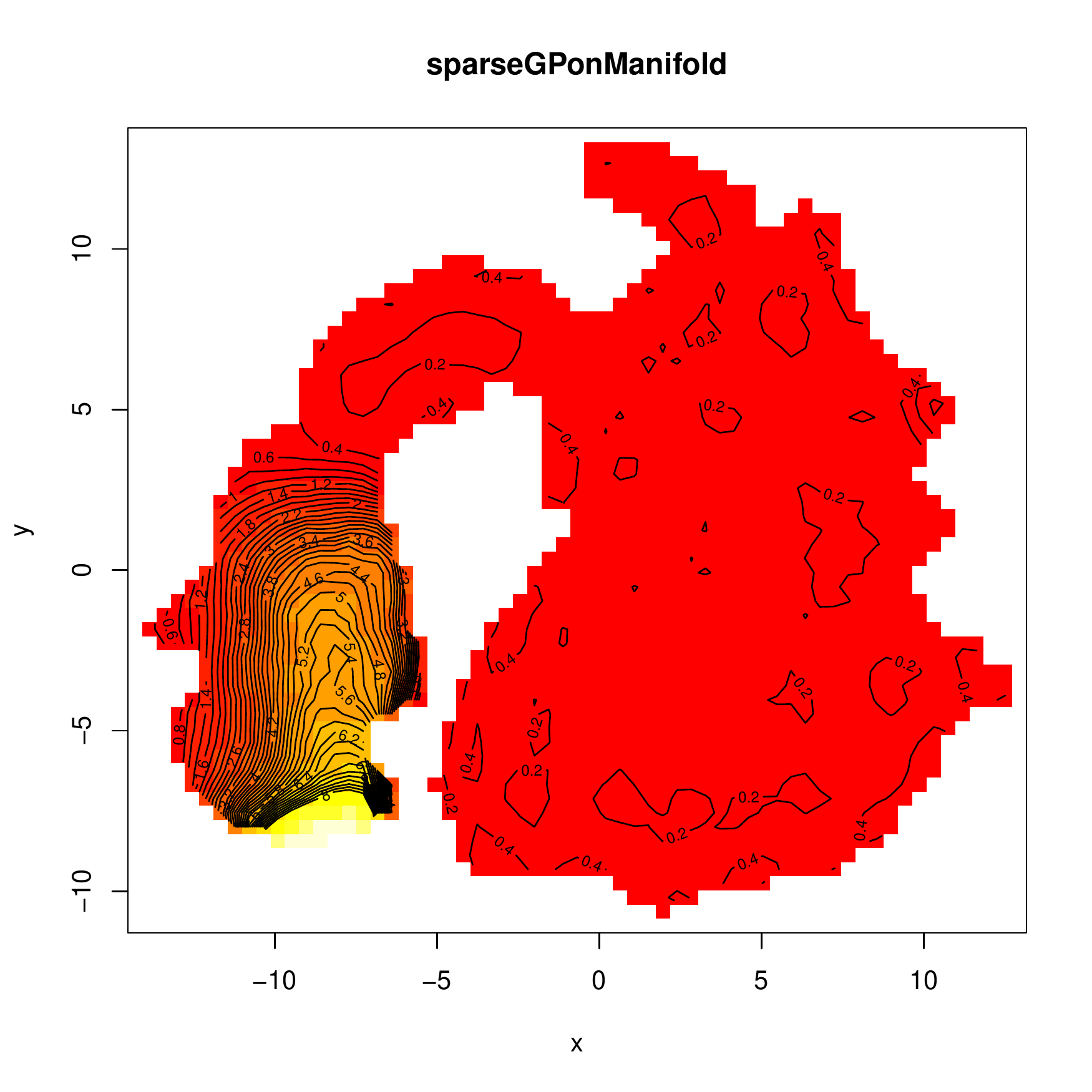} }
\footnotesize
    \caption{ \label{fig:example}
   Predictive variance of in-GP on grid points. }
\end{figure}

The predictive mean of the normal GP is shown in Figure \ref{aral:gp}. As expected, the normal GP  smoothes across the isthmus of the central peninsula. Relatively high levels of chlorophyll concentration are estimated for the southern part of the eastern shore of the western basin of the sea, while all observations in this region have rather low concentrations. Similarly a decline in chlorophyll level towards the southern half of the western shore of the eastern basin is estimated, which is different from the pattern of the data in the region.    On the other hand, the predictive mean using sparse in-GP  does not produce these artefacts (see Figure \ref{aral:gpmani}) and tracks the data pattern better. 
The value of predictive variance is plotted as a heat map in Figure \ref{fig:var}.

These artefacts become even more serious when the coverage of the data is uneven. In Figure \ref{aral:datal} we removed most of the data points in the southern part of the western basin of the sea, and  the same models are applied to this uneven dataset.  Figure \ref{aral:gpl} shows the normal GP extrapolation across the isthmus from the eastern basin of the sea. In contrast, the sparse in-GP estimates  as plotted in Figure \ref{aral:gpmanil} do not seem to be affected by the data from the eastern side of the isthmus. The value of predictive variance is plotted as a heat map in Figure \ref{fig:varleft}. Since most of the data points in the southern part of the western basin of the sea have been removed, the values of the variance estimates have increased towards the southern end of the western sea.

%\begin{landscape}
%\end{landscape}

\section{Discussion}

Our work proposes a novel  class of intrinsic Gaussian processes on manifolds and complex constrained domains employing the equivalence relationship between  heat kernels  and  the transition density of  Brownian motions on manifolds.  One of the key features of in-GP is to fully incorporate  the intrinsic geometry of the spaces  for inference while respecting the potential complex boundary constraints or interiors.  To reduce the computational cost of simulating BM sample paths when the sample size is large, sparse in-GPs are developed leveraging ideas from the literature on fast computation in GPs in Euclidean
spaces.  The results in section \ref{simdata} and \ref{real} indicate that in-GP achieves significant improvement over usual GPs.

The focus of this article has been on developing in-GPs on manifolds with known metric tensors.  There has been abundant interest in learning of unknown lower-dimensional subspace or manifold structure in high-dimensional data.  Our method can be combined with these approaches for performing supervised learning on lower-dimensional latent manifolds.

\section*{Acknowledgement}
LL  acknowledge support for this article from NSF grants DMS  CAREER 1654579 and IIS 1663870.

\appendix

\section{Choosing the sample and window sizes in the case of  $\bb{R}$}
In this appendix, we illustrate our methodology by providing details of its application to estimating the heat kernel of $\mathbb{R}$.  Since the expression of the heat kernel is known in this case, the estimation error can be measured by simulating different number of BM sample paths.  

Consider a BM $\{S(t):t>0\}$ in $\bb{R}$ with $S(0)=0$. Fix some $t>0$ and $s\in\bb{R}$.
The probability density of $S(t)$ at $s$ is 
\begin{align}
\label{eqn:trueR}
K =  \frac{1}{\sqrt{2\pi t}} \exp \left( -\frac{s^2}{2t} \right) = K_{heat}^t(0,s)
\end{align}
which is also the heat kernel in $\bb{R}$. Our method of estimating the transition probability $K$ consists of two parts.% $\hat{p}'$

%\begin{itemize}
%\item
{\bf Part 1}. Choose a window size $w$ and approximate $K$ by
\begin{align}
\label{eqn:real1}
K'=\frac{1}{2w}\,\bb{P}\big[|S(t)-s|<w\big] = \frac{1}{2w} \int_{s-w}^{s+w} \frac{1}{\sqrt{2\pi t} } exp\left( -\frac{y^2}{2t} \right) dy.
\end{align}
Assuming that $w$ is small compared to $\sqrt{t}$, by taking the Taylor expansion of equation \ref{eqn:real1}  we have
\begin{align}
\label{eqn:tay}
K'=K+\frac{K(s^2-t)}{6t}\cdot\frac{w^2}{t}
  +O\left(\frac{w^4}{t^2}\right).
\end{align}
In particular, the approximation error increases (quadratically) with $w$.

{\bf Part 2}. Then choose a sample size $N$ and estimate $K'$ by 
\begin{align}
\hat{K}=\frac{1}{2w}\cdot\frac{k}{N},\quad
\t{where }k\sim\t{Bin}(N,2wK').
\end{align}
Recall that $k$ is the number of sample paths with $|S(t)-s|<w$ and it has binomial distribution with number of trial as $N$ and probability of success as $2wK'$.
The standard error of this estimator is
\begin{align}
\label{eqn:bin}
\frac{1}{2Nw}\cdot\sqrt{N\cdot 2wK'(1-2wK')}
\leq\sqrt{\frac{K'}{2Nw}}
\approx\sqrt{\frac{K}{2Nw}}
\end{align}
which decreases with $w$.
%\end{itemize}

Under the assumption that $w\ll \sqrt{t}$, the window size $w$ is optimal when the sum of the two errors described in equation \eqref{eqn:tay} and equation \eqref{eqn:bin} is minimal.
\begin{align}
\hat{K} - K \approx \left( \frac{K(s^2-t)}{6t} \cdot \frac{w^2}{t} \right)+ \sqrt{ \frac{K}{2Nw} }
\end{align}
The optimal order of magnitude of $w$ can be calculated by minimising $(\hat{K}-K)^2$. This optimal window size and the corresponding total error are of the following orders of magnitude:
\begin{align}
w_{\t{opt}}\sim A^{-2/5}K^{-1/5}t^{2/5}N^{-1/5},\label{eqn:wop} \qquad
\t{err}\sim A^{1/5}K^{3/5}t^{-1/5}N^{-2/5},
\end{align}
where $A=|s^2-t|/t$. The sample size $N$ has to be large enough to guarantee that $w_{\t{opt}}\ll \sqrt{t}$ and at the same time the total error is within a predetermined desired level.
This turns out to be that
\begin{align}
\label{eqn:N}
N\gg A^{-2}K^{-1}t^{-1/2}\quad\t{and}\quad
N\gtrsim A^{1/2}K^{-1}t^{-1/2}(\t{err}/K)^{-5/2}.
\end{align}
Given a predefined error $err$, $A^{1/2}K^{-1}t^{-1/2}(\t{err}/K)^{-5/2}$ can be seen as the minimum number of BM sample paths simulation required. 

\section{Choosing the sample and window sizes in the case of  $\bb{R}^2$.}

\noindent Consider a Brownian motion $\{X(t):t>0\}$ in $\bb{R}^2$ with $X(0)=(0,0)$.
Fix some $t>0$ and $(x,y)\in\bb{R}^2$.
The probability density of $X(t)$ at $(x,y)$ is
\begin{align*}
K = \frac{1}{2\pi t}\exp\left(-\frac{x^2+y^2}{2t}\right) = K_{heat}^t  \big( ( 0,0 ), (x,y) \big)
\end{align*}
which is the same as the value of the heat kernel $K_t$ of $\bb{R}^2$ at $((0,0),(x,y))$.
Our method of estimating $K$ consists of two parts.

{\bf Part 1}. First choose a window size $w$ and approximate $p$ by
\begin{align*}
K'=\frac{1}{4w^2}\,\bb{P}\big[\,||X(t)-(x,y)||<w\,\big].
\end{align*}
Assuming that $w$ is small compared to $\sqrt{t}$, we have
\begin{align*}
K'=K+K\frac{(x^2+y^2-2t)}{6t}\cdot\frac{w^2}{t}
  +O\left(\frac{w^4}{t^2}\right).
\end{align*}
In particular, the approximation error increases (quadratically) with $w$.

{\bf Part 2}. Then choose a sample size $N$ and estimate $K'$ by 
\begin{align*}
\hat{K}'=\frac{1}{4w^2}\cdot\frac{k}{N},\quad
\t{where }k\sim\t{Bin}(N,4w^2 K').
\end{align*}
(Recall that $k$ is the number of sample paths with $||X(t)-(x,y)||<w$.)
The standard error of this estimator is
\begin{align*}
\frac{1}{4Nw^2}\cdot\sqrt{N\cdot 4w^2 K'(1-4w^2 K')}
\leq\sqrt{\frac{K'}{4Nw^2}}
\approx\sqrt{\frac{K}{4Nw^2}}
\end{align*}
which decreases with $w$.

Under the assumption that $w\ll \sqrt{t}$, the window size $w$ is optimal when the two errors described above are similar in magnitude.
This optimal window size and the corresponding total error are of the following orders of magnitude:
\begin{align*}
w_{\t{opt}}\sim A^{-1/3}K^{-1/6}t^{1/3}N^{-1/6},\qquad
\t{err}\sim A^{1/3}K^{2/3}t^{-1/3}N^{-1/3},
\end{align*}
where $A=|x^2+y^2-2t|/t$.
Meanwhile, the sample size $N$ has to be large enough to guarantee that $w_{\t{opt}}\ll \sqrt{t}$ and at the same time the total error is within a predetermined desired level.
This turns out to mean that
\begin{align*}
N\gg A^{-2}K^{-1}t^{-1}\quad\t{and}\quad
N\gtrsim A^{1/3}K^{-1/3}t^{-1/3}(\t{err}/K)^{-1}.
\end{align*}

\section{Brownian motion on Swiss Roll.}

The three-dimensional coordinates of Swiss Roll can be parametrized by two variables, namely $r$ the radius and $z$ the width. Consider the Swiss roll parametrized by
\begin{align*}
\mathbf{x}(r,z)=(r\cos r,r\sin r,z).
\end{align*}
To find its metric tensor, we first compute the partial derivatives
\begin{align*}
\mathbf{x}_r &= (\cos r-r\sin r, \sin r+r\cos r, 0) \\
\mathbf{x}_z &= (0,0,1)
\end{align*}
The metric tensor is given by
\begin{align*}
&\quad(\mathbf{x}_r\cdot\mathbf{x}_r)dr^2
+2(\mathbf{x}_r\cdot\mathbf{x}_r)dr\,dz
+(\mathbf{x}_z\cdot\mathbf{x}_z)dz^2 \\
&=(1+r^2)dr^2+dz^2
\end{align*}
or in matrix form
\begin{align*}
g= \left[\begin{array}{cc}  
1+r^2 & 0 \\[0.3em]
0 & 1 \\[0.3em]
           \end{array} \right], \qquad
 g^{-1}= \left[\begin{array}{cc} 
\frac{1}{1+r^2} & 0 \\[0.3em]
0 & 1 \\[0.3em]
           \end{array} \right], \qquad
\frac{\partial g}{\partial r}= \left[\begin{array}{cc}  
2r &0 \\[0.3em]
 0 & 0 \\[0.3em]
                \end{array} \right]            
\end{align*}

The general equation of BM on manifold is given as
\begin{align*}
dx_i(t) = \sum^{2}_{j=1} \left(  -g^{-1} \frac{\partial g}{\partial x_j(t)} g^{-1} \right)_{ij} dt + \frac{1}{2} \sum_{j=1}^2(g^{-1})_{ij}tr(g^{-1}\frac{\partial g}{\partial x_j(t)})dt + \left( g^{-1/2} dB(t)\right)_i
\end{align*}
where $G$ is determinant of metric tensor and $B(t)$ is an independent BM in Euclidean space. Substituting $g$ into above equation, the BM on the Swiss Roll can be written as
\begin{align}
\label{eqn:swissBM}
dr(t) &= \left( -g^{-1} \frac{\partial g}{\partial r} g^{-1} \right )_{11} dt + \frac{1}{2} g^{-1}_{11} tr( g^{-1} \frac{\partial g}{\partial r} )  dt + (g^{-1/2})_{11}dB_r  \\
dr(t) &= \frac{-2r}{(1+r^2)^2} dt + \frac{1}{2} \frac{2r}{(1+r^2)^2} dt + (1+r^2)^{-1/2} dB_r(t) \nonumber  \\
dz(t) &=  (g^{-1/2})_{22}dB_z(t) \\
dz(t) &=  dB_z(t)  \nonumber
\end{align}

%\newpage

%\begin{algorithm}[!h]
%\caption{Deciding if a point $(x,y)$ lies inside or outside a polygon with vertices, in order, $(a_1,b_1),(a_2,b_2),\ldots,(a_n,b_n)$}
%
%\begin{algorithmic}[1]
%
%\State $\theta\gets 0$\qquad
%\Comment{$\theta$ will measure the angle swept by a line connecting $(x,y)$ and the boundary, as the boundary is traversed}
%
%\State $(a_{n+1},b_{n+1})\gets (a_1,b_1)$
%
%\For{$i=1,2,\ldots,n$}
%
%\State $U\gets (a_i-x)(a_{i+1}-x)+(b_i-y)(b_{i+1}-y)$
%
%\State $V\gets (a_i-x)(b_{i+1}-y)-(b_i-y)(a_{i+1}-x)$
%
%\If{$V\geq 0$}
%
%\State $d\theta\gets \cos^{-1}\big(U/\sqrt{U^2+V^2}\big)$, with value between $0$ and $\pi$
%
%\Else
%
%\State $d\theta\gets (-1)\cdot\cos^{-1}\big(U/\sqrt{U^2+V^2}\big)$, with value between $-\pi$ and $0$
%
%\EndIf
%
%\State $\theta\gets \theta+d\theta$
%
%\EndFor
%
%\State Conclusion: If $\theta=0$, it means $(x,y)$ is outside the polygon. If $\theta=2\pi$ or $-2\pi$, it means $(x,y)$ is inside the polygon. There is no other possibility.
%
%\end{algorithmic}
%\end{algorithm}
%

\bibliography{heatkern}
\bibliographystyle{rss}

\end{document}